\documentclass{article}

\usepackage[preprint]{neurips_2026}

\usepackage[utf8]{inputenc} % allow utf-8 input
\usepackage[T1]{fontenc}    % use 8-bit T1 fonts
\usepackage{hyperref}       % hyperlinks
\usepackage{url}            % simple URL typesetting
\usepackage{booktabs}       % professional-quality tables
\usepackage{amsfonts}       % blackboard math symbols
\usepackage{nicefrac}       % compact symbols for 1/2, etc.
\usepackage{microtype}      % microtypography
\usepackage{xcolor}         % colors
\usepackage{wrapfig}

\usepackage{amsmath}
\usepackage{amssymb}
\usepackage{mathtools}
\usepackage{amsthm}
\usepackage{multirow}
\usepackage{pifont}
\usepackage{threeparttable}
\usepackage{soul}
\usepackage{enumitem}
\usepackage{tcolorbox}
\usepackage{subcaption}
\theoremstyle{plain}

\theoremstyle{definition}

\theoremstyle{remark}

\newtheorem{observation}{Observation}
\usepackage{algorithm}
\usepackage{algpseudocode}

\usepackage{xcolor}
\usepackage{forest}
\usepackage{minted}
\title{Are LLMs Ready for Neural-integrated Mechanistic Modeling? A Benchmark and Agentic Framework}

\author{%
  Zihan Guan$^{*,1}$, Rituparna Datta$^{*,1}$, Mengxuan Hu$^{1}$, Shunshun Liu$^{1}$, \\ \textbf{Aiying Zhang}$^{1}$,  \textbf{Prasanna Balachandran}$^{1}$, \textbf{Sheng Li}$^{1}$, \textbf{Anil Vullikanti}$^{1}$ \\
  $^{*}$ Equal Contributions, $^{1}$University of Virginia
}

\begin{document}

\maketitle

\begin{abstract}
Large language models (LLMs) have shown promise in constructing mechanistic models from data. However, existing evaluations largely focus on simplified settings and fail to capture the complexity of real-world scientific modeling. In practice, such modeling often involves neural-integrated formulations, where a mechanistic model component and a neural network component are jointly constructed, leading to a significantly more complex search space. Motivated by this gap, we introduce the Neural-Integrated Mechanistic Modeling (NIMM) benchmark, which evaluates LLM-generated neural-integrated mechanistic models across three scientific domains. Experiments on NIMM reveal that existing LLM-based approaches struggle to effectively explore this complex space, resulting in limited search stability and solution quality. To address this challenge, we propose NIMMGen, a tree-guided agentic framework that enables diversified exploration via branch-level search and improves solutions through atomic model refinement. Extensive experiments demonstrate that NIMMGen achieves state-of-the-art performance on NIMM, significantly improving search stability and solution quality.
\end{abstract}

\section{Introduction}

Mechanistic models, constructed based on domain knowledge of physical, biological, or chemical processes, provide a principled framework for understanding and predicting complex systems and have been widely used across scientific domains, including medicine~\cite{marchal2025applications, katsoulakis2024digital}, power systems~\cite{meyur2022ensembles, thorve2023high}, and computational epidemiology~\cite{aseam-survey, marathe:cacm13, cui2025identifying, chopra2022differentiable}. However, constructing such models typically requires substantial domain expertise and manual effort, making the process time-consuming and difficult to scale.
Motivated by recent advances in large language models (LLMs) for code generation and agentic optimization~\cite{novikov2025alphaevolve, tang2024biocoder, white2023assessment}, a growing line of work~\cite{holt2024automatically, amad2025continuously} has begun to explore LLM-driven frameworks that learn and refine mechanistic models directly from data.

Among various forms of mechanistic models, neural-integrated mechanistic models~\cite{chopra2022differentiable, nunez2023forecasting, ning2023epi, guan2025framework, cui2025identifying, datta2025calypso, liu2022integrating}, which augment mechanistic formulations with neural components, have recently gained significant attention. These models achieve stronger performance by enabling the representation of complex system dynamics. However, this added flexibility also introduces substantial challenges, as both the model structure and neural components must be \textit{jointly constructed and refined}, leading to a significantly more complex modeling and search space.

However, existing evaluation environments in the LLM-based approaches~\cite{holt2024automatically, amad2025continuously} are overly simplified and fail to capture the complexity of neural-integrated modeling. In particular, these works either focus on purely mechanistic models or restrict hybrid models to narrowly defined forms (e.g., additive combinations of neural and mechanistic components), which represent only a limited subset of the broader neural-integrated modeling space. Therefore, this gap limits progress toward automated scientific model discovery, where effective mechanistic models must be constructed for complex real-world scientific systems.

To address this gap, we introduce the \textit{Neural-Integrated Mechanistic Modeling} (NIMM) benchmark, which is designed to systematically evaluate whether LLMs can construct and refine neural-integrated mechanistic models under realistic scientific settings. We use datasets from three diversified domains, including public health, clinical health, and materials science. In contrast to prior work, NIMM additionally incorporates key characteristics of real-world modeling: \textit{partial observability}. For example, in epidemiological datasets, only reported infection counts are observed, while latent compartments such as susceptible or exposed populations remain unobserved. 
% \anil{we dont have partial observation in the cancer and material science applications}
We evaluate a range of baselines, including black-box neural networks, existing mechanistic models, and LLM-driven approaches, under the NIMM framework. Our results show that current LLM-based methods exhibit limited search stability and solution quality when constructing neural-integrated mechanistic models, as evidenced by high root mean squared error (RMSE) and low execution success rates (ESR).

% Guided by these observations, we propose \textbf{NIMMGen}, a tree-guided agentic framework for neural-integrated mechanistic modeling. NIMMGen promotes exploration diversity through structured branch-level search, while improving stability via fine-grained, atomic model refinements.
Guided by these observations, we propose \textbf{NIMMGen}, a tree-structured agentic framework for neural-integrated mechanistic modeling. NIMMGen organizes model construction as a branching search process, where planning, model generation, execution feedback, and refinement are orchestrated along the search tree. This structure promotes exploration diversity through branch-level search while improving stability via fine-grained, atomic model refinements. Empirically, NIMMGen achieves up to \textbf{95.1\%} reduction in RMSE on the public health subset, \textbf{92.6\%} reduction in RMSE on the clinical health subset, and \textbf{24.5\%} on the materials science subset, compared to prior SOTA LLM-based baselines (Table~\ref{tab:evaluation_results}). In terms of stability, NIMMGen further improves executability, yielding up to \textbf{76.8\%} higher ESR on the public health subset, \textbf{20.8\%} higher ESR on the clinical health subset, and \textbf{18.9\%} on the materials science subset (Table~\ref{tab:evaluation_results}). Beyond predictive accuracy, we demonstrate that the learned models support counterfactual intervention analysis (e.g., social distancing policies), highlighting their practical utility for decision-making (Figure~\ref{fig:intervention}). 
% \anil{Counter-factual analyses require the models to be correct.
% Also we dont do counter-factual analyses for the other two datasets.
% }
Further analysis shows that the tree structure helps the LLM progressively refine the model (Figure~\ref{fig:loss_iteration}) and introduce meaningful structural updates that improve model adaptability over iterations (Figure~\ref{fig:tree}).

In summary, this paper makes three contributions:

\textbf{(1) Neural-integrated mechanistic modeling as a new evaluation setting.} We identify neural-integrated mechanistic modeling as an important yet underexplored setting for evaluating LLM-based mechanistic modeling, where neural and mechanistic components are jointly constructed. To study this setting, we introduce NIMM, a benchmark that systematically evaluates model construction and refinement under realistic constraints; \textbf{(2) Revealing Limitations from the Existing LLM-based approaches.} We show that current methods struggle in this setting, exhibiting unstable search behavior and low-quality solutions, highlighting fundamental challenges in neural-integrated mechanistic modeling; \textbf{(3) A tree-guided agentic framework for model construction.} We propose NIMMGen, a tree-guided framework that improves neural-integrated model generation through structured exploration and fine-grained refinement, achieving state-of-the-art performance.

\section{Background}
Let $\mathcal{S} := (\mathcal{X}, \mathcal{U}, \Phi)$ denote a dynamical system, which consists of $d_{\mathcal{X}}$-dimensional states from its state space $\mathcal{X} \subseteq \mathbb{R}^{d_\mathcal{X}}$, the (optional) $d_{\mathcal{U}}$-dimensional action space $\mathcal{U} \subseteq \mathbb{R}^{d_\mathcal{U}}$, and a dynamical model $\Phi$ (following notation from \cite{holt2024automatically}). 
At a given time $t \in \mathcal{T} \subseteq \mathbb{R}_{+}$, the state of the system is denoted as $x(t) \in \mathcal{X}$, while the action of the system is $u(t) \in \mathcal{U}$. In continuous time, the evolution of the system $\mathcal{S}$ is given by
\begin{equation}
    \frac{d{x(t)}}{d_t} = \Phi(x(t), u(t), t)
\end{equation}
with $\Phi : \mathcal{X} \times \mathcal{U} \times \mathcal{T} \rightarrow \mathcal{X}$.

For example, in an epidemic dynamic system, the state of the system $x(t)$ may represent the number of susceptible, infected, and recovered individuals.
% \anil{explain the terms in an example, such as epidemic models}

\textbf{Mechanistic Models.} 
% \anil{could drop digital twins. 
% These seem to be coming all of a sudden.
% We could just say we are using $f$ to approximate $\Phi$, which is not known exactly
% }
We usually approximate $\Phi$ using a computational model $f_{\theta} \in \mathcal{F}$ informed by data, where $\theta \in \Omega(\theta)$ denotes the parameters of the model. 
Here, the sets $\mathcal{F}, \Omega(\theta)$ denote the space of candidate model classes and parameter space, respectively. The parameters $\theta$ are usually calibrated manually~\cite{kemp2021modelling, davids2020sabcom} or informed from data using statistical methods~\cite{anderson2001ensemble, yang2015inference}. 

In scientific domains, the system dynamics is not purely statistical but constructed grounded with domain knowledge derived from known physical, biological, or chemical processes~\cite{marchal2025applications, katsoulakis2024digital, meyur2022ensembles, cui2025identifying}. This structure requires each component in $f$ to have a clear interpretation and supports reasoning about interventions and counterfactuals.
% \anil{should revise the above paragraphs slightly, so this text doesnt look like a copy of \cite{holt2024automatically}.
% Example is useful to explain the dynamical system
% }

\textbf{Neural-integrated Mechanistic Models.}
Recent work~\cite{chopra2022differentiable, nunez2023forecasting, ning2023epi, guan2025framework, cui2025identifying, datta2025calypso, liu2022integrating} has shown that black-box neural networks can be effective for calibrating mechanistic models. Specifically, let $f^{NN}_{\beta}: \mathcal{D} \rightarrow \Omega({\theta})$ be a neural network parameterized by $\beta$ that maps the input data domain $\mathcal{D}$ to the mechanistic model parameters space.  
% \anil{$\beta$ not clear}
The resulting mechanistic model is then written as $f^{mech}_{\theta}$, where the parameter vector $\theta$ is directly provided by the $f^{NN}_{\beta}$, i.e., 
\begin{equation}
    \theta = f^{NN}_{\beta}(D), D \in \mathcal{D}.
\end{equation}

This yields a compositional model $f$ in which the neural network predicts the parameters for the mechanistic component:
\begin{equation}
    f(D) = f^{mech} \circ f^{NN} (D) = f^{mech}({f_{\beta}^{NN}(D)}).
\end{equation}

\begin{figure}
    \centering
    \includegraphics[width=\linewidth]{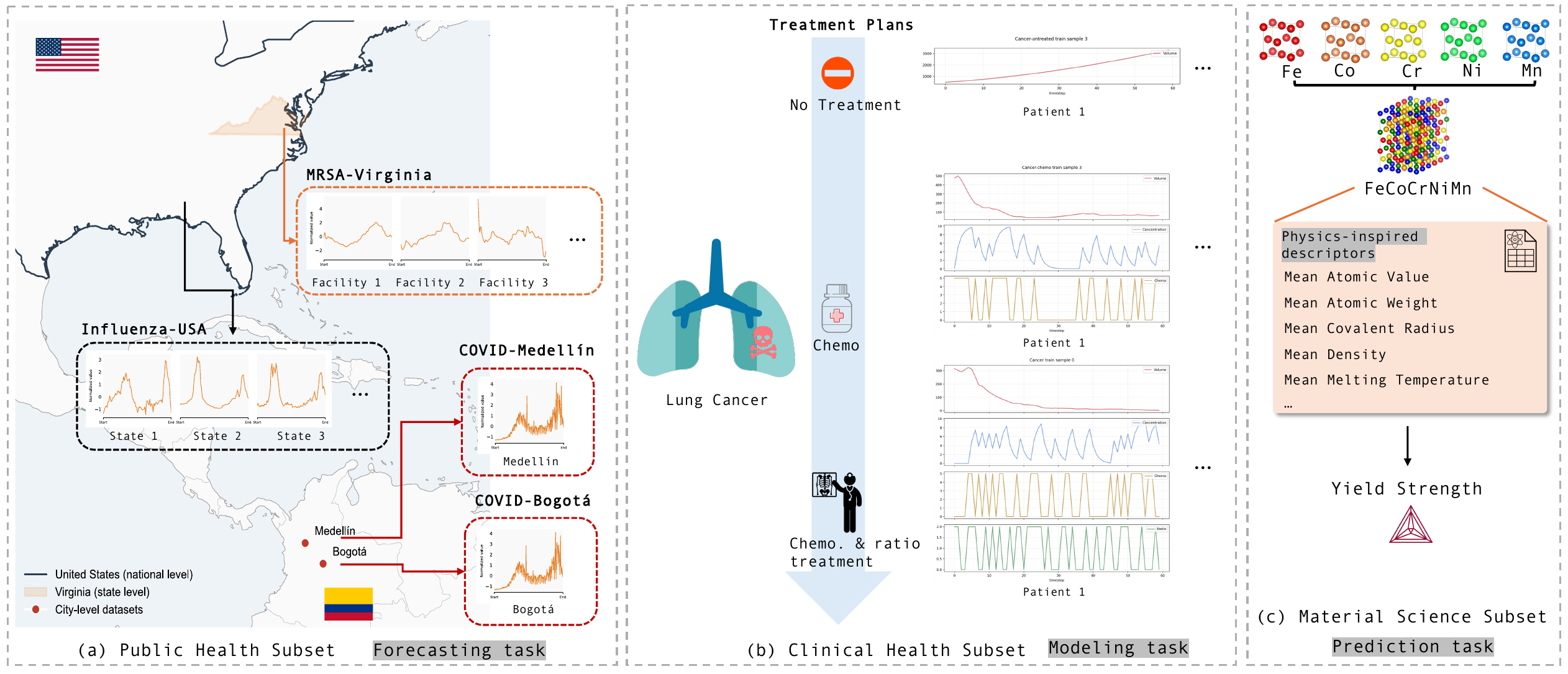}
    \caption{Overview of the NIMM benchmark across three domains: (a) Public Health, (b) Clinical Health, and (c) Materials Science. The benchmark includes epidemic forecasting tasks, lung cancer modeling tasks, and alloy yield strength prediction. NIMM evaluates whether LLMs can construct effective \textit{neural-integrated mechanistic models} across diversified scientific tasks.
    % \ritu{In the fig, Material Science should be (c)}
    }
    \label{fig:dataset}
    \vspace{-5mm}
\end{figure}

\section{The NIMM benchmark.}

% \anil{still not clear about "benchmark". Doesnt seem consistent with usage in other papers. Could just call it NIMM framework and move the discussion above this (explaining the "Neural-integrated Mechanistic Models" here)}

\subsection{Motivation}\label{sec:motivation}
Despite preliminary efforts to apply LLMs to mechanistic modeling~\cite{holt2024automatically, amad2025continuously}, we argue that their evaluation environment is oversimplified and cannot capture the complexity of neural-integrated modeling. Specifically, they either focus on purely mechanistic models or assume that the target model $f$ takes the form of an \textit{additive combination} of a neural network and a mechanistic component, which is a special case of the general neural-integrated mechanistic model we consider in the paper. 

Apart from that, we additionally argue that the exisiting evaluation environments fail to align with the real-world scientific modeling: \ding{182} Unlike the existing problem setting~\cite{holt2024automatically, amad2025continuously}, where mechanistic models are used under \emph{full observation} (i.e., all $d_\mathcal{X}$-dimensional states are observable in the data), a more realistic setting is \emph{partial observation}, where only a subset of the states are observed.  For example, in epidemiological datasets, only reported infection counts are observed, while latent compartments, such as susceptible or exposed populations, cannot be easily observed by the surveillance system. \ding{183} In many scientific applications, the goal is not merely to learn mechanistic models, but also to apply these models for diversified tasks. 
A desired problem setting should therefore evaluate models in diversified scenarios, rather than focusing on modeling in isolation.

These considerations motivate the need for a benchmark that formalizes the new problem setting and systematically evaluates an LLM’s ability to perform neural-integrated mechanistic modeling.

\subsection{NIMM benchmark}
\textbf{Datasets.} Neural-integrated mechanistic modeling has been widely applied across a variety of domains, including public health~\cite{chopra2022differentiable, guan2025framework, datta2025calypso, cui2025identifying}, clinical health~\cite{geng2017prediction, holt2024automatically}, and materials science~\cite{liu2022integrating}. Accordingly, in the NIMM benchmark, we include datasets from all three domains. An overview of the datasets is shown in Figure~\ref{fig:dataset}.

The three subsets have different tasks (\underline{Point \ding{183}}). In this section, we describe the \textbf{Public Health} subset. The \textbf{Clinical Health} and \textbf{Materials Science} subsets are described in detail in Appendix~\ref{appendix:cancer} and Appendix~\ref{appendix:materials}, respectively. For the \textbf{Public Health} domain, we use four widely used datasets from prior work: influenza data from the United States (Influenza-USA), methicillin-resistant \textit{Staphylococcus aureus} data from Virginia (MRSA-Virginia), and COVID-19 data from two South American cities—Bogota (COVID-Bogota) and Medellin (COVID-Medellin).

% We also use datasets from other domains, including \textbf{Clinical Health} and \textbf{Materials Science}, which are described in detail in \S~\ref{sec:experimental_setting}.  

% Figure~\ref{fig:dataset} visualizes the collected datasets. Each dataset is spatiotemporal, e.g., MRSA-Virginia is a weekly dataset collected from 644 healthcare facilities in Virginia from 2016-01-03 to 2021-01-24. Figure~\ref{fig:dataset} visualizes the geolocations and time series of each dataset. In this section, we present the descriptions for the \textbf{Public Health} subset.

\textbf{Task Formulation.} For public health, we adopt spatial-temporal forecasting as the main task for evaluating the effectiveness of the generated mechanistic models, which is one of the most popular tasks in epidemics. Accordingly, we consider $D$ as a spatial-temporal dataset, where $D_{l, t}$ denotes the signals for the location $l \in \{1,...,L\}$ at timestamp $t$. The predicted parameters $\theta$ are also \textit{spatially-} and \textit{time-varying}, where $\theta_{l,t}$ denotes the predicted mechanistic parameter for the location $l$ at the timestamp $t$. This fine-grained formulation enables the mechanistic model to better adapt to heterogeneous dynamics across both space and time.  We also note that the signals $D_{l,t} \in \mathbb{R}^d$ may be multivariate, containing a \textit{subset} of the $d_\mathcal{X}$-dimensional states (e.g., infection counts) together with other heterogeneous contextual features (\underline{Point \ding{182}}).

% \anil{the problems we handle are not just time series now, so would change this to more general version}
Formally, the problem goal is to use data till time $T$, namely $\{{D}_{:,t}\}_{t=1}^T$,
to predict $\{{\hat{y}}_{:, t}\}_{t=T+1}^{T+H}$, e.g.,
the number of infections and deaths during the time period $[T+1, T+H]$, where $H$ is the forecast horizon. More specifically, the historical data $\{{D}_{:, t}\}_{t=1}^T$ is used to train the compositional model $f$ as in Equation~\ref{eqn:train}. In the inference stage, given the historical data, the neural network component $f^{NN}$ first generates spatially- and time-varying parameters $\theta_{{l, t}}, l \in \{1,...,L\}, t\in\{1,...,T+H\}$. Then the mechanistic component $f^{mech}$ is run $T+H$ steps under these predicted parameters, where the last $H$ simulation steps serve as our forecasts.

\textbf{Neural-integrated Setups.} In a typical neural-integrated mechanistic modeling pipeline, both a neural component $f^{\mathrm{NN}}$ and a mechanistic component $f^{\mathrm{mech}}$ are required. 
To evaluate the LLM's capabilities across different practical settings, we define the following scenarios: 
(1) \textbf{Mechanistic Mode}: completing a mechanistic model when the neural network component is already provided in the environment; 
(2) \textbf{Hybrid Mode}: jointly completing both the mechanistic model and the neural network model. 
The hybrid mode is intuitively more challenging than the mechanistic mode, as it involves a substantially larger search space.
% \anil{instead of benchmarks, I would say these as additional tasks we are interested in}

\textbf{Metrics.} We aim to evaluate the 
\textit{effectiveness} and \textit{executability} of the generated model. For effectiveness, we employ popular metrics such as root mean square error (RMSE) for the forecasting task. Formally, the RMSE is defined as \(\sqrt{\frac{1}{L \cdot H}\sum_{l,t=T+1}^{t=T+H} \|\hat{y}_{l, t} - y_{l,t}\|^2}\).
Following the prior work~\cite{chopra2022differentiable}, we adopt a more practical but challenging \textit{real-time} evaluation setup, where the training and testing period shifts over time, to avoid the cherry-picking fallacy. More details about the setup can be found in the Appendix~\ref{appendix:task}. For executability evaluations, we use the execution success rate (ESR)~\cite{chen2025pcebench} as the metric, where the ESR evaluates the percentage of code samples that not only compile but also execute without runtime errors. Formally, ESR is defined as \(\frac{\text{\# Exectuable Codes}}{\text{Total \# Generations}} \times 100\).

\begin{figure*}[!t]
    \centering
    \includegraphics[width=\linewidth]{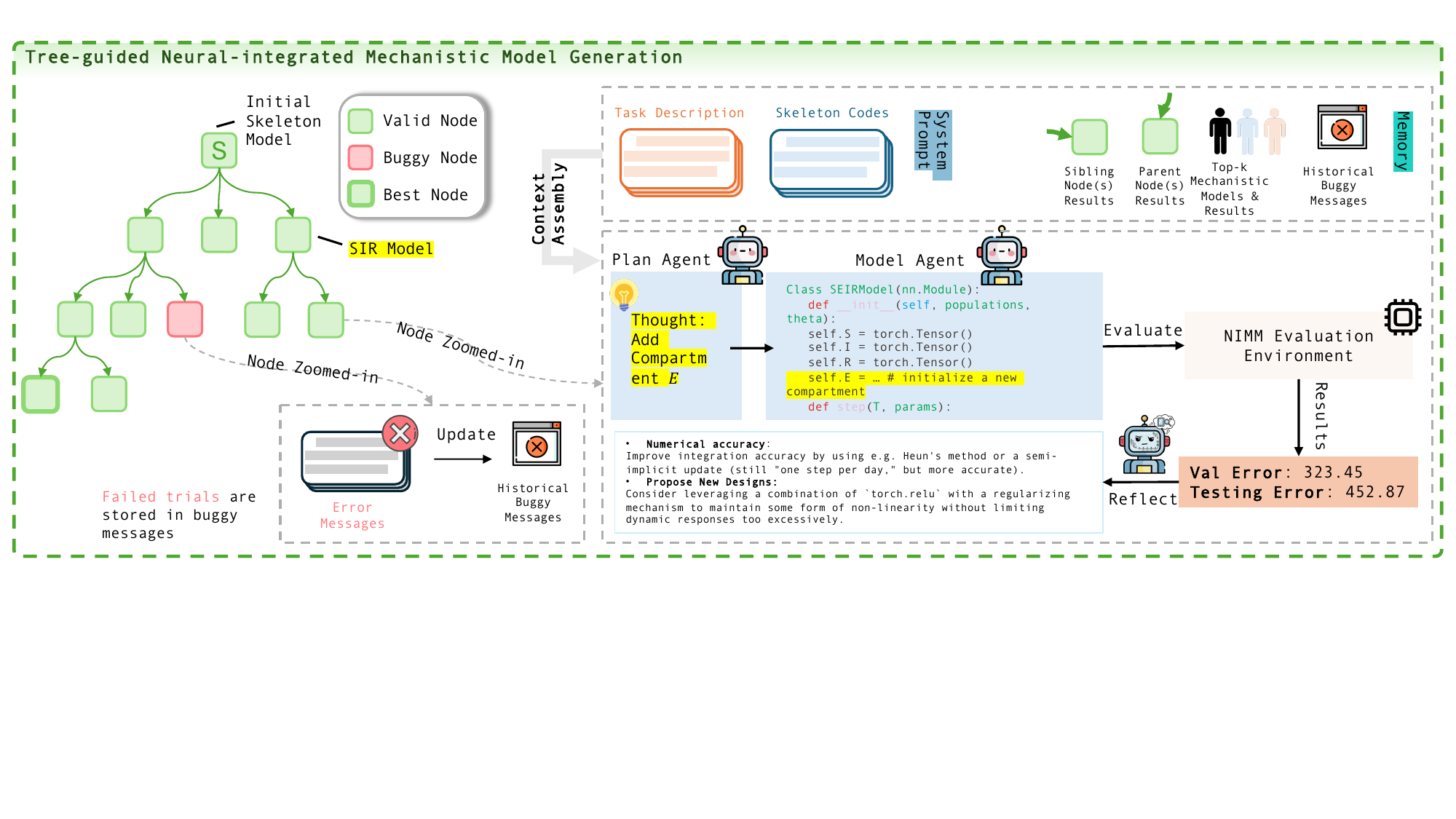}
    \caption{Overview of the NIMMGen framework. The initialized skeleton model serves as the root node, and the model is iteratively refined through a search tree. For each valid node, NIMMGen uses the assembled contextual information to perform three steps: model generation by a model agent, model evaluation in the NIMM environment, and result reflection by a reflection agent. For buggy nodes whose execution fails, the resulting error message is stored to update the contextual memory.}
    \label{fig:pipeline}
    \vspace{-2mm}
\end{figure*}

\textbf{Baselines.} The LLM-based baselines include zero-shot prompting and HDTwinGen~\cite{holt2024automatically}. In addition, we consider purely black-box models such as Transformers~\cite{vaswani2017attention} and LSTMs~\cite{hochreiter1997long}. We also include prior neural–integrated mechanistic models for comparison, including Grad-Metapopulation~\cite{guan2025framework}.
% \anil{we can say that baselines for the other tasks are described later.
% But metrics and baselines seem out of place here.
% They can be put in the experiments section}

\begin{table*}[!t]
    \centering
    \caption{Comparison of the effectiveness and executability in the NIMM benchmark.}
    \label{tab:evaluation_results}
    \begin{threeparttable}
    \resizebox{\textwidth}{!}{\begin{tabular}{c|cccccccccccc}
    \toprule
       \multirow{3}*{Dataset $\downarrow$} & \multirow{3}*{Mode} & {LSTM} & {Transformer} & {Grad-Metapopulation} & \multicolumn{2}{c}{Zero-shot} & \multicolumn{2}{c}{HDTwinGen} & \multicolumn{2}{c}{Ours}\\
       \cmidrule(lr){3-3} \cmidrule(lr){4-4} \cmidrule(lr){5-5} \cmidrule(lr){6-7} \cmidrule(lr){8-9} \cmidrule(lr){10-11}
        & & RMSE $\downarrow$ & RMSE $\downarrow$ & RMSE $\downarrow$ & RMSE $\downarrow$ & ESR\tnote{\#} $\uparrow$  & RMSE $\downarrow$ & ESR $\uparrow$ & RMSE $\downarrow$ & ESR $\uparrow$\\
       \midrule
        \multirow{2}*{COVID-Bogota} & Mechanistic &  \multirow{2}*{$633.88_{\pm 158.06}$} & \multirow{2}*{$901.96_{\pm 284.58}$} & \multirow{2}*{$1362.89_{\pm 165.26}$} & $1102.62_{\pm 476.82}$ & $0.80$ & $339.13_{\pm 86.53}$ & $0.69_{\pm 0.18}$ & $\textbf{325.22}_{\pm 82.52}$ & $\textbf{0.98}_{\pm 0.01}$ \\
        & Hybrid & & & & $1358.24_{\pm 266.25}$ & $0.40$ & $765.27_{\pm 64.58}$ & $0.57_{\pm 0.13}$ & $\textbf{398.62}_{\pm 131.57}$ & $\textbf{0.98}_{\pm 0.03}$ \\
        \midrule
        \multirow{2}*{COVID-Medellin} & Mechanistic & \multirow{2}*{$586.88_{\pm 16.96}$} & \multirow{2}*{$552.39_{\pm 46.89}$} & \multirow{2}*{$743.25_{\pm 25.79}$} & $908.19_{\pm 357.17}$ & $0.80$ & $652.72_{\pm 102.79}$ & $0.62_{\pm 0.10}$ & ${\textbf{480.51}}_{\pm 172.47}$ & $\textbf{0.85}_{\pm 0.09}$ \\
        & Hybrid & & & & $1026.25_{\pm 262.64}$ & $0.75$ & $675.82_{\pm 190.89}$ & $0.80_{\pm 0.13}$ &  $\textbf{529.65}_{\pm 207.86}$ & $\textbf{0.84}_{\pm 0.11}$\\
        \midrule
        \multirow{2}*{Influenza-USA} & Mechanistic & \multirow{2}*{$86.69_{\pm 1.73}$} &  \multirow{2}*{$30.72_{\pm 0.27}$} & \multirow{2}*{$117.90_{\pm 10.93}$} & $20.92_{\pm 5.38}$ & $0.80$ & $8.78_{\pm 1.49}$ & $0.89_{\pm 0.11}$ & $\textbf{7.57}_{\pm 2.05}$ & $\textbf{0.98}_{\pm 0.01}$ \\
        & Hybrid & & & & $783.49_{\pm 1240.58}$ & $0.70$ & $163.69_{\pm 40.75}$ & $0.60_{\pm 0.08}$ & $\textbf{8.09}_{\pm{3.27}}$ & $\textbf{0.89}_{\pm 0.13}$ \\
        \midrule
        \multirow{2}*{MRSA-Virginia} & Mechanistic & \multirow{2}*{$167.08_{\pm 8.52}$} & \multirow{2}*{$127.53_{\pm 45.59}$} & \multirow{2}*{$134.05_{\pm 16.29}$} & $516.36_{\pm 646.61}$ & $0.70$ & $93.23_{\pm 5.19}$ & $0.76_{\pm 0.04}$ & $\textbf{33.74}_{\pm 0.46}$ & $\textbf{0.99}_{\pm 0.01}$\\
        & Hybrid & & & &  $809.99_{\pm 302.54}$ & $0.65$ & $132.67_{\pm 42.82}$ & $0.56_{\pm 6.44}$ & $\textbf{38.58}_{\pm 14.62}$ & $\textbf{0.99}_{\pm 0.01}$\\
        \midrule
        \midrule
        \multirow{2}*{Cancer (NT)} & Mechanistic & \multirow{2}*{$3.48_{\pm 1.46}$} & \multirow{2}*{$4.24_{\pm 1.74}$} & \multirow{2}*{/} & $73.25_{\pm 45.1}$ & $0.70$  & $1.69_{\pm 2.05}$ & $0.73_{\pm 0.04}$ & $\textbf{0.22}_{\pm 0.01}$ & $\textbf{0.88}_{\pm 0.04}$\\
        & Hybrid & & & & $89.65_{\pm 28.72}$ & $0.70$ & $1.97_{\pm 1.62}$ & $0.74_{\pm 0.06}$ & $\textbf{0.35}_{\pm 0.17}$ & $\textbf{0.89}_{\pm 0.11}$\\
        \midrule
        \multirow{2}*{Cancer (w/ C)} & Mechanistic & \multirow{2}*{$0.62_{\pm 0.17}$} & \multirow{2}*{$0.58_{\pm 0.29}$} & \multirow{2}*{/} & $53.52_{\pm 16.22}$ & $0.65$ & $1.23_{\pm 0.04}$ & $0.77_{\pm 0.04}$ & $\textbf{0.10}_{\pm 0.05}$ & $\textbf{0.93}_{\pm 0.03}$  \\
        & Hybrid & & & & $69.87_{\pm 5.82}$ & $0.65$ & $1.63_{\pm 0.62}$ & $0.72_{\pm 0.06}$ & $\textbf{0.12}_{\pm 0.01}$ & $\textbf{0.85}_{\pm 0.07}$\\
        \midrule
        \multirow{2}*{Cancer (w/ C \& R)} & Mechanistic & \multirow{2}*{$0.93_{\pm 0.34}$} & \multirow{2}*{$1.76_{\pm 0.28}$} & \multirow{2}*{/} & $109.25_{\pm 25.64}$ & $0.70$ & $0.78_{\pm 0.46}$ & $0.82_{\pm 0.12}$ & $\textbf{0.32}_{\pm 0.11}$& $\textbf{0.95}_{\pm 0.04}$ \\
        & Hybrid & & & & $129.20_{\pm 15.97} $ & $0.65$ & $1.47_{\pm 0.64}$ & $0.79_{\pm 0.03}$ & $\textbf{0.23}_{\pm 0.17}$ & $\textbf{0.89}_{\pm 0.06}$\\
        \midrule
        \midrule
        \multirow{2}*{Materials (FCC)} & Mechanistic & \multirow{2}*{$213.13_{\pm 58.51}$}\tnote{*} & \multirow{2}*{/} & \multirow{2}*{/} & $268.24_{\pm 12.90}$ & 0.75 & $192.67_{\pm 11.30}$ & $0.74_{\pm 0.07}$ & $\textbf{145.43}_{\pm 25.14}$ & $\textbf{0.85}_{\pm 0.03}$\\
        & Hybrid & & & & $274.30_{\pm 16.72}$ & 0.65 & $208.66_{\pm 25.62}$ & $0.80_{\pm 0.06}$ & $\textbf{176.04}_{\pm 10.52}$ & $\textbf{0.87}_{\pm 0.02}$\\
        \midrule
        \multirow{2}*{Materials (BCC)} & Mechanistic & \multirow{2}*{$235.10_{\pm 15.55}$}\tnote{*} & \multirow{2}*{/} &\multirow{2}*{/} & $267.84_{\pm 14.42}$ & $0.65$ & $192.65_{\pm 10.51}$ & $0.79_{\pm 0.04}$ & $\textbf{166.01}_{\pm 29.37}$ & $\textbf{0.89}_{\pm 0.04}$  \\
        & Hybrid & & & & $277.17_{\pm 74.21}$ & $0.80$ & $212.65_{\pm 14.35}$ & $0.74_{\pm 0.03}$ & $\textbf{189.92}_{\pm 17.25}$ & $\textbf{0.88}_{\pm 0.01}$\\
    \bottomrule
    \end{tabular}}
    \begin{tablenotes}
    \tiny
    \item[*] We use an MLP-based neural network rather than LSTM for the Materials task.
    \item[\#] The metric is intentionally abused. We report the ESR obtained by running the zero-shot method 20 times.
    \end{tablenotes}
    \end{threeparttable}
    \vspace{-5mm}
\end{table*}
  
% \begin{remark}[Dual Role of the NIMM benchmark]
%     The NIMM benchmark not only introduced a new setting to evaluate LLMs’ ability to generate code under practical scenarios, but also serves as a probe of LLM's potential performance in scientific domains.
% \end{remark}

\subsection{Evaluation Results}
Table~\ref{tab:evaluation_results} reports the results of the baseline methods on the NIMM benchmark. We run each method three times and report the average RMSE values together with the ESR. Our results reveal several key observations.
\begin{tcolorbox}
\begin{observation}[\textbf{Search Stability}]
Existing LLM-based approaches frequently generate mechanistic models that fail due to run-time errors.
\label{obs:high_bug}
\end{observation}
\end{tcolorbox}

The average ESR values for the zero-shot and the HDTwinGen are 0.69 and 0.73, respectively. For the zero-shot method, the low ESR primarily stems from the difficulty of the NIMM tasks: the LLM may fail to fully understand the task requirements and generate runnable models in a single attempt. For HDTwinGen, we hypothesize that the low ESR arises from the combination of its sequential search process and its coarse-grained model revision strategy. On the one hand, HDTwinGen adopts a sequential refinement strategy, in which each candidate is generated by holistically revising models from the previous round. This makes the search highly dependent on a single trajectory, implicitly allowing early errors or suboptimal structures to propagate across iterations. On the other hand, HDTwinGen might unexpectedly employ large and entangled revisions, making the search difficult to control, as failures cannot be easily attributed to a specific change, and useful components may be unintentionally broken.
% \anil{all comparison is with HDTwinGen.
% Are there any other subsequent methods to say something about?}

\begin{tcolorbox}
\begin{observation}[\textbf{Solution Quality}]
Existing LLM-based approaches fail to generate effective models that fit well with the NIMM evaluation environments.
\label{obs:real-time}
\end{observation}
\end{tcolorbox}

Even among successful trials, the resulting models often still exhibit high RMSE. For example, across the nine evaluated datasets, HDTwinGen yields higher RMSE than deep learning–based approaches (e.g., LSTM) in five cases under the hybrid mode. This suggests that the search process is not only unstable but also limited in its capacity to discover accurate mechanistic structures. We argue that this weakness also arises from the reliance on a single search trajectory, which restricts the diversity of exploration.

% \begin{observation}
% LLMs might generate mathematically valid but semantically incorrect models.
% \label{obs:wrong_codes}
% \end{observation}

% In these baselines, the primary objective is to generate mechanistic models that fit the observed data and perform forecasting through simulation. However, optimizing solely for this objective can lead LLMs to produce mechanistic models that are mathematically valid yet semantically wrong. 

% These models yield no run-time errors, but the specifications can hardly be interpreted by the human evaluators. In Appendix~\ref{appendix:semantical_wrong_codes}, we present qualitative examples of such generated semantically wrong models.

% \anil{the discussion for these observations could be tightened}

\section{Methodology}
\subsection{Overview}
This section presents NIMMGen, a tree-guided agentic framework, designed to address the two observations above. In contrast to existing sequential search strategies, we adopt a tree-guided search strategy~\cite{lu2026towards, yao2023tree} that combines branch-level exploration with atomic model refinement. The branching structure prevents flawed intermediate solutions from dominating the entire search process while also increasing search diversity. Atomic edits make each refinement step more localized and controllable. Together, these design choices reduce error propagation and increase search diversity, leading to higher ESR and lower RMSE. An overview of the NIMMGen is provided in Figure~\ref{fig:pipeline}.

% Specifically, we introduce (i) a self-evolving loop consisting of a modeling agent and a reflection agent, which progressively refines model specifications through iterative optimization, addressing the challenges in Observation~\ref{obs:real-time}; (ii) an error-handling module that augments the agent’s memory with historical error messages to reduce repeated failures, addressing Observation~\ref{obs:high_bug}; 
% and (iii) a verification agent that evaluates the physical and semantic consistency of generated models, addressing Observation~\ref{obs:wrong_codes}. 

Given the system prompt $C$, NIMMGen performs $G \in \mathbb{N}^{+}$ iterations of optimization under a tree-guided search framework. The search starts from an initialized model, which serves as the root node of the search tree. Each node in the tree corresponds to a concrete candidate neural-integrated mechanistic model, and the edges correspond to the atomic refinement applied to the model. In this formulation, a child node is obtained by applying a structural modification (e.g., adding a new compartment or transition) to its parent model.

Instead of sequentially refining a single trajectory, NIMMGen iteratively expands valid nodes in the tree, enabling branch-level exploration of multiple models. For each selected node, the framework assembles contextual information from the task description, skeleton codes, system prompt, parent and sibling node results, top-$k$ historical mechanistic models, and stored buggy messages. Conditioned on this context, a plan agent first proposes an atomic refinement action (e.g., adding a new compartment $E$ to the parent model), and a model agent implements the corresponding structural modification to produce a new candidate model.
% \anil{not clear what are parent-child relationships in models, and how these are being expanded}

The generated model is then evaluated in the NIMM environment, and the resulting feedback is summarized by a reflection agent to update the search memory. If the candidate is buggy, its error message is recorded in the buggy-memory bank, which would be revisited only if there are no valid expandable nodes in the search tree. By combining branch-level exploration with atomic refinement, NIMMGen makes the search process more controllable, preserves diversity across candidate trajectories, and reduces error propagation during model generation.

% \textbf{Code RAG Database.} To further enrich the agent’s context, we introduce an external code RAG database $D_{code}$ that allows it to proactively retrieve useful code snippets. $D_{code}$ contains the code snippets collected from the relevant literature in the past five years. Formally, the code snippets are extracted by $E^g = h(D_{code}, R^{g-1})$, where the reflection at the last step $R^{g-1}$ is used as the query prompt, and the code snippets $E^g \subset D_{code}$ will be returned. Detailed setups are provided in Appendix~\ref{appendix:rag}.

\subsection{Main Loop}
\textbf{System Prompt.}
The system prompt $C$ consists of two components: (1) $T$, a task description in natural language that specifies the input and output variables as well as the modeling objective; and (2) $S$, skeleton code that defines the template of the neural-integrated mechanistic model.
The skeleton code $S$ provides essential structural information, such as function signatures and required code formats, which guides the generation of executable model implementations in a predefined format. For illustration, examples of the skeleton code are provided in Appendix~\ref{appendix:prompts}.

\textbf{Tree-guided Search.} Starting from a single model $f^0$ initialized from the skeleton $S$, the framework builds a search tree by iteratively expanding one parent node at a time. At iteration $g$, NIMMGen first constructs an expandable set under depth, child-capacity, and validation-quality constraints. Formally,

\begin{equation}
      \mathcal{F}^{g}_{\mathrm{valid}}
      =
      \left\{
      v \in \mathcal{T}^{g-1}_{\mathrm{valid}}
      :
      d(v)<D_{\max},\;
      c(v)<C_{\max},\;
      L_{\mathrm{val}}(v)\le \tau_{\mathrm{val}}
      \right\}
\end{equation}

where $\mathcal{T}^{g-1}_{\mathrm{valid}}$ denotes the valid nodes in the tree at step $g-1$, $d(v)$ denotes the depth of the node $v$, $c(v)$ denotes the child capacity of the node $v$, and  $L_{\mathrm{val}}(v)$ denotes the validation loss evaluated with the model in node $v$.
% \anil{not clear what a node is and what the children are}

Then, we select the node to expand by lexicographically minimizing the following tuple, where the priority is given in the listed order:

\[
  v_p^g=
  \arg\min_{v\in\mathcal{F}_{\mathrm{valid}}^g}
  \Bigl(
  L_{\mathrm{val}}(v),\;
  -d(v),\;
  c(v),\;
  \mathrm{id}(v)
  \Bigr),
  \]

\textbf{Context Assembly.}
At each generation step $g$, NIMMGen assembles contextual information from multiple sources. Specifically, the context includes: (1) the task description and skeleton code in the system prompt; (2) the parent node result and sibling node results associated with the currently expanded node $v$; (3) a population of top-$k$ historical mechanistic models, along with their evaluation results; and (4) historical error messages from previously failed models.
Together, these sources provide rich contextual signals for the agent to reason over prior successes and failures, support branch-level exploration, and guide subsequent model refinement.

\noindent
\textbf{Plan Agent.}
Before model generation, the plan agent proposes a localized refinement plan for the selected node based on the assembled context. Rather than making a holistic revision to the entire model, the plan agent identifies an atomic modification, such as adding a new term, revising an equation, or adjusting a dependency between variables. In Figure~\ref{fig:pipeline}, the plan agent proposes adding an exposed compartment $E$ to the parent susceptible-infected-recovered ($SIR$) model, yielding a susceptible-exposed-infected-recovered ($SEIR$) model. This step makes the search process more structured and controllable by refining the model step-by-step with localized modifications.

\noindent
\textbf{Modeling Agent.}
Conditioned on the system prompt $C$, the assembled context at step $g$, and the refinement plan proposed by the plan agent, the modeling agent generates a new neural-integrated mechanistic model:
\[
f = f^{mech}(f_{\beta}^{NN}) \sim \texttt{LLM}_{m}(C, M^g),
\]
where $M^g$ denotes the assembled context associated with the selected tree node.

\noindent
\textbf{Model Evaluation.}
The generated model interacts with the NIMM environment for evaluation. Specifically, the objective is to fit the neural-integrated model to the training data, so as to obtain the optimized parameter set $\beta^*$. The learnable hyperparameters may vary across tasks; following prior work~\cite{holt2024automatically}, we provide the full details in the Appendix.

\noindent
\textbf{Reflection Agent.}
After evaluation, the reflection agent summarizes the feedback for the newly generated candidate from multiple perspectives, including both model specification and implementation quality. These reflections are written in natural language and added back to memory to guide future node expansions. Formally, the reflection process can be written as
\[
R^g \leftarrow \texttt{LLM}_{r}(v^g),
\]
where $v^g$ denotes the evaluation feedback at step $g$.

\noindent
\textbf{Buggy Nodes.} If the generated candidate fails to execute, its error message is stored in the buggy-memory bank, and the corresponding branch is terminated rather than further expanded. After completing the $G$ iterations, NIMMGen selects the best-performing model across all explored nodes as the final output. The buggy nodes can only be revisited for debugging if there are no valid nodes in the expandable set.

Due to the space limit, more detailed descriptions of the NIMMGen algorithm are in the Appendix~\ref{appendix:algorithm}.

% \subsection{Environment Analysis Module}\label{sec:insight}
% Apart from the above designs, we implement an environment analysis module at the initial stage of our pipeline to better adapt the agent to the environment. Specifically, we enable the agent to actively perceive environmental signals by enriching its context with raw data, system information, and library versions. The raw data allows the agent to directly extract insights and propose targeted designs for simulation. These insights are then added back into the context as references when the modeling agent generates model code. Including system and library version information helps the agent anticipate compatibility issues, avoid version-related errors, and understand which tools and data structures are available in advance.

\section{Evaluation}
\subsection{Main Evaluations}\label{sec:experimental_setting}
\textbf{Experimental Setup.}
To demonstrate the effectiveness of the NIMMGen, we evaluate its performance on the three domains in the NIMM benchmark. The iteration step is chosen as 40, $D_{\max}$ and $C_{\max}$ are chosen as 6, by default. All these hyperparameters are evaluated with sensitivity studies later.

% \begin{table}[!t]
%     \centering
%     \caption{Comparison of NIMMGen with the baselines on the Cancer dataset.}
%     \label{tab:cancer}
%     \resizebox{\linewidth}{!}{\begin{tabular}{c|ccc}
%         \toprule
%          & Zero Shot & HDTwinGen  & Ours \\
%          \midrule
%          Lung Cancer & $73.25_{\pm 45.1}$ & $1.69_{\pm 2.05}$ & $1.41_{\pm 1.58}$\\
%          Lung Cancer (w/ Chemo.) & $53.52_{\pm 16.22}$ & $1.23_{\pm 0.04}$ & $0.08_{\pm 0.04}$\\
%          Lung Cancer (w/ Chemo. \& Radio.) & $109.25_{\pm 25.64}$ & $0.78_{\pm 0.46}$ & $0.06_{\pm 0.02}$\\
%          \bottomrule
%     \end{tabular}}
%     \vspace{-5mm}
% \end{table}

\textbf{Effectiveness.}
Table~\ref{tab:evaluation_results} reports the performance comparison between the NIMMGen and the baselines. Overall, the NIMMGen achieves substantially lower RMSE across all datasets. Compared to the existing LLM-based baselines such as zero-shot prompting and HDTwinGen, NIMMGen achieves up to \textbf{95.1\%} reduction in RMSE on the public health subset, \textbf{92.6\%} reduction in RMSE on the clinical health subset, and \textbf{24.5\%} on the materials science subset.

\textbf{Searching Stability.} In terms of stability, NIMMGen further improves executability of the constructed models, yielding up to \textbf{76.8\%} higher ESR on the public health subset, \textbf{20.8\%} higher ESR on the clinical health subset, and \textbf{18.9\%} on the materials science subset.

\textbf{Comparison between the Mechanistic and the Hybrid Models.} We also observe that the hybrid mode generally shows a slightly worse performance than the mechanistic mode. Although this may appear counterintuitive, it aligns with our expectations: In the hybrid mode, the LLM is expected to tune both the mechanistic part and the neural network part, which might be harder to synchronize. The challenge also comes from the fact that the searching space of the hybrid space is considerably larger compared to the mechanistic mode. Therefore, given the same budget, the hybrid mode shows a slightly worse performance.

\subsection{Further Analysis}\label{sec:analysis}
\begin{wrapfigure}{r}{0.45\textwidth}
    \centering
    \vspace{-4mm}
    \includegraphics[width=0.45\textwidth]{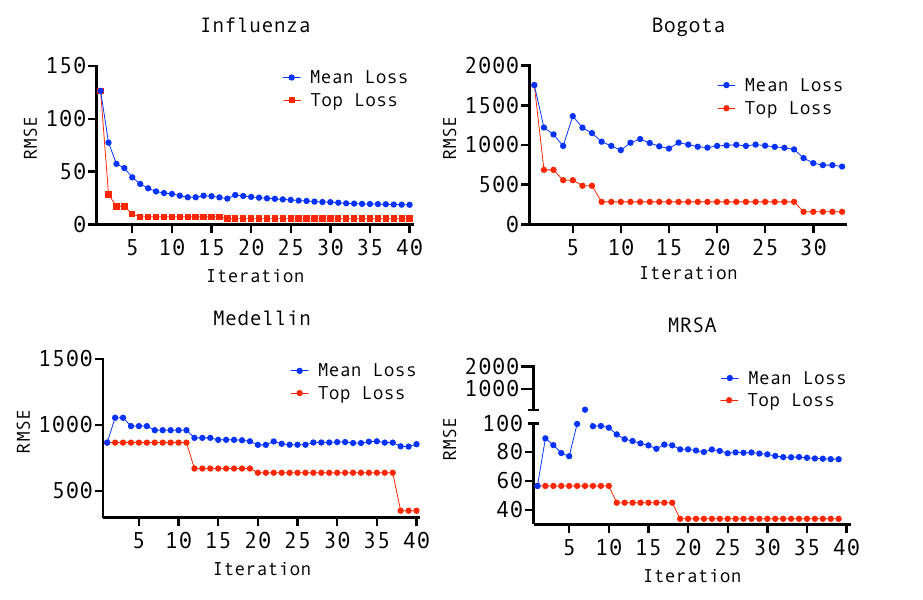}
    \caption{Loss curve over iterations.}
    \label{fig:loss_iteration}
    \vspace{-4mm}
\end{wrapfigure}

\textbf{What has been changed during the Optimization?}
Figure~\ref{fig:loss_iteration} visualizes the loss curve over the course of optimization. We plot both the average RMSE validation loss of all historically generated mechanistic models and the best (lowest) RMSE validation loss achieved so far. As shown, the mean loss exhibits a decreasing trend, suggesting that models generated at later iterations progressively improve the population through guided optimization rather than purely stochastic trial-and-error. The best loss also steadily decreases with iterations, demonstrating that the NIMMGen can continuously refine the model during the search process. 

\textbf{Case Study of the Expansion Tree.} Figure~\ref{fig:tree} shows an example of the expansion tree in a typical run in the Influenza-USA dataset. As shown, the expansion tree reveals that NIMMGen explores multiple candidate models through branch-level search, where each node corresponds to a refined model and is evaluated by its validation loss. The best-performing model is reached through a sequence of iterative refinements from the root. To further understand what changes in the mechanistic code contribute to these improvements, we qualitatively analyze the mechanistic models generated by LLMs throughout the path to the best node, with the key modifications summarized in Table~\ref{tab:influenza_search_summary}.

\begin{wraptable}{r}{0.58\textwidth}
    \centering
    \caption{Performance comparison under different LLMs.}
    \label{tab:model_usage}
    % \vspace{-2mm}
    \resizebox{0.56\textwidth}{!}{%
    \begin{tabular}{l|lcc}
        \toprule
        Model Type & Model Name & Influenza-USA & MRSA-Virginia \\
        \midrule
        \multirow{3}*{Proprietary LLM}
        & \includegraphics[width=0.02\textwidth]{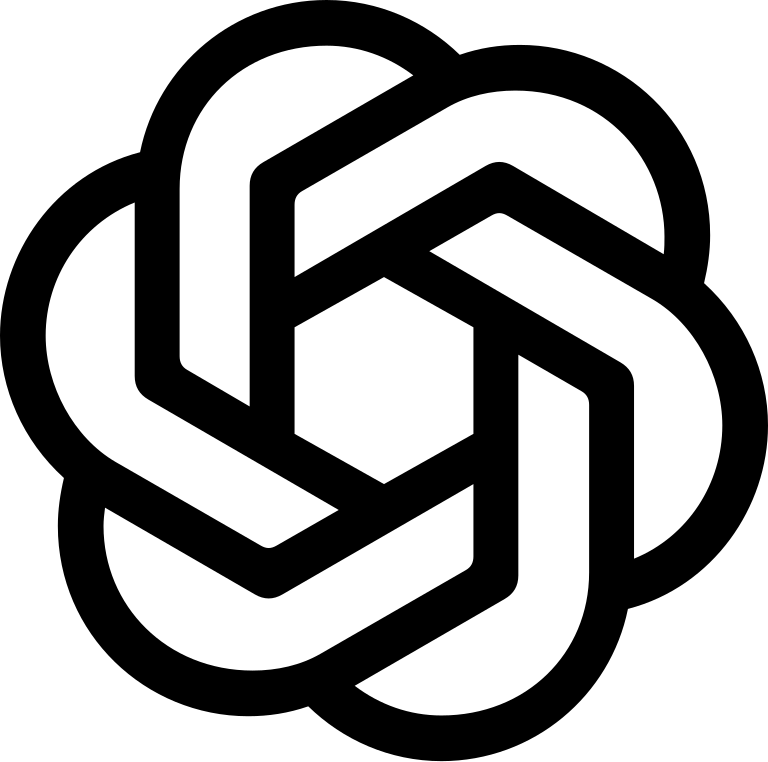} \texttt{GPT-4.1} 
        & $7.57_{\pm 2.05}$ & $33.74_{\pm 0.46}$ \\
        & \includegraphics[width=0.02\textwidth]{Figures/openai_logo.png} \texttt{GPT-4o} 
        & $7.20_{\pm 1.56}$ & $38.15_{\pm 5.25}$ \\
        & \includegraphics[width=0.02\textwidth]{Figures/openai_logo.png} \texttt{GPT-5-mini} 
        & $10.20_{\pm 0.37}$ & $39.12_{\pm 1.07}$ \\
        \midrule
        \multirow{2}*{Open-source LLM}
        & \includegraphics[width=0.02\textwidth]{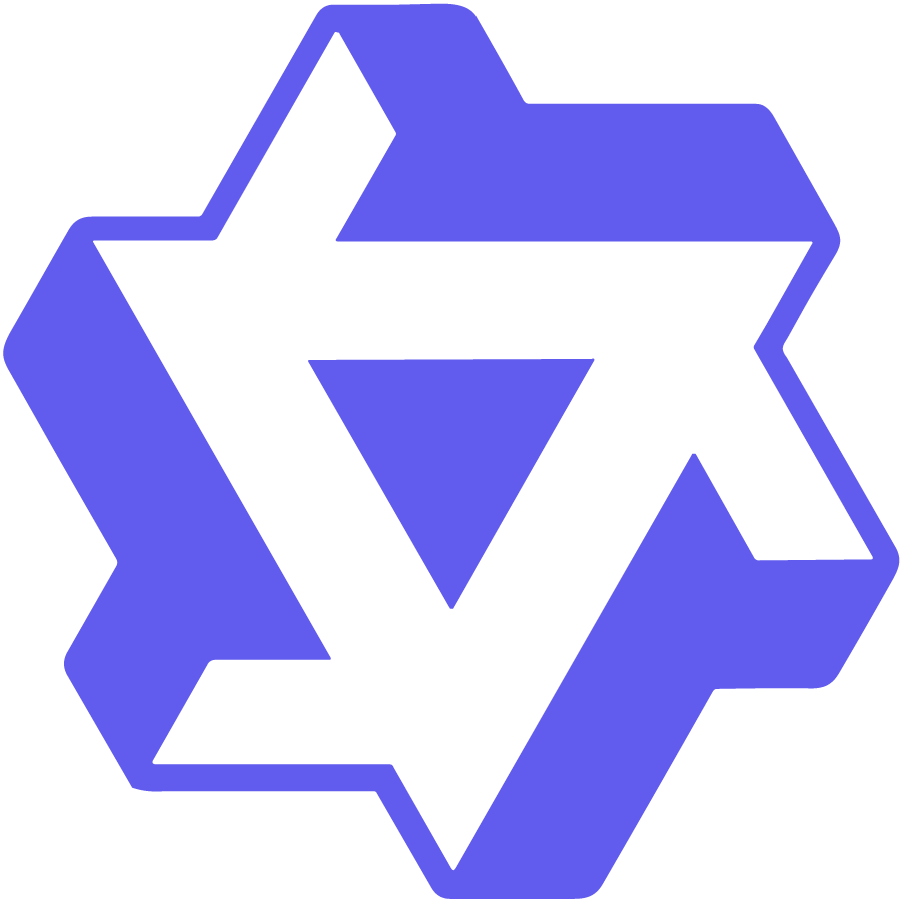} \texttt{Qwen2.5-32B-Coder} 
        & $24.26_{\pm 3.12}$ & $57.45_{\pm 12.98}$ \\
        & \includegraphics[width=0.02\textwidth]{Figures/qwen_logo.png} \texttt{Qwen3-30B-Coder} 
        & $20.24_{\pm 10.08}$ & $58.25_{\pm 16.91}$ \\
        \bottomrule
    \end{tabular}}
    % \vspace{-4mm}
\end{wraptable}

\textbf{Sensitivity of the Hyper-parameters.}
In Table~\ref{tab:model_usage}, we compare the performance of the NIMMGen with different base LLMs, including proprietary models such as \texttt{GPT-4.1}, \texttt{GPT-4o}, and \texttt{GPT-5-mini}, and two frontier open-source coding models, including \texttt{Qwen2.5-32B-Coder} and \texttt{Qwen3-30B-Coder}. The detailed setups are provided in Appendix~\ref{appendix:open-source}. In Table~\ref{tab:iteration}, we compare the performance of the NIMMGen under different iteration numbers, ranging from 10 to 80. In Figure~\ref{fig:heatmap}, we compare the performance of the NIMMGen with different $D_{max}$ and $C_{max}$ choices.

\begin{wrapfigure}{r}{0.4\textwidth}
    \centering
    \vspace{-7mm}
    \includegraphics[width=0.4\textwidth]{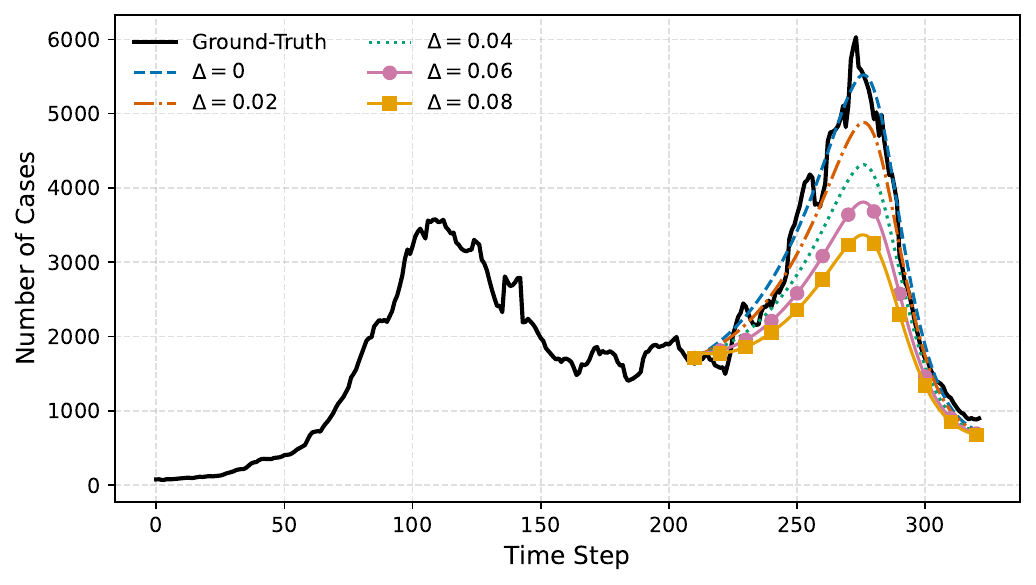}
    \caption{Observed and simulated epidemic trajectories under social distancing interventions.}
    \label{fig:intervention}
    \vspace{-7mm}
\end{wrapfigure}

\subsection{Counterfactual Intervention Simulations} \label{sec:application}
In this section, we show the potential validity of the models generated by NIMMGen with a counterfactual intervention simulation experiment.

Counterfactual intervention simulation is a widely used application of mechanistic models, aiming to quantify how hypothetical policies (e.g., social distancing) alter epidemic trajectories relative to the observed baseline~\cite{ge2021impact, rǎdulescu2020management, ventura2022modeling}. In particular, social distancing policies primarily affect disease transmission by reducing effective contact rates. We therefore simulate social distancing by adjusting the transmission-related component of the model after a given time $T'$. Let $\theta_{l,t,\bar{\gamma}}$ denote the effective transmission rates $\bar{\gamma}$ produced by the neural component $f^{NN}_{\beta}$ at time $t$ for location $l$. We define
\begin{equation}
    \theta_{l,t,\bar{\gamma}} = 
\begin{cases}
\theta_{l,t,\bar{\gamma}}, & t < T'\\
(1-\Delta)\cdot \theta_{l,t,\bar{\gamma}}, & t \ge T'
\end{cases}, \forall 1 < l \leq L
\end{equation}

where $\Delta$ denotes the strength of social distancing intervention and $\Delta$ a larger $\Delta$ corresponds to stronger reductions in population contact.
% \anil{parameters not defined earlier. Need some clarification}

Figure~\ref{fig:intervention} shows the intervention experiments on the COVID-Bogota dataset ($L=1$) with $T'=270$. The neural-integrated mechanistic model is chosen as the best one generated by the NIMMGen. As expected, increasing the strength of the simulated social distancing intervention results in systematic reductions in epidemic peak magnitude and cumulative case counts. 
% \anil{see the setting we considered in the PET paper.
% It was connected to some event that happened in Colombia at that time. Should refer to that}
\ul{Notably, this experiment does not aim to assess the real-world effectiveness of specific policies, as counterfactual ground truth is inherently unavailable.} Instead, the objective is to \ul{evaluate whether the generated mechanistic models respond to intervention parameters in a manner consistent with epidemiological principles.}

\subsection{Limitations and Future Works}
Despite the great effectiveness we show in the evaluation section, there are several main limitations that can be improved in future work: (1) The definition of the atomic operation is still coarse-grained. A more interesting direction is to use an additional ontology-guided knowledge graph to construct the domain-specific atomic sets. This might potentially further enhance the agentic system with the domain knowledge; (2) We mainly focus on the effectiveness (i.e., RMSE) of these models. Although we show the potential validity of the constructed model by using the counterfactual analysis experiment, we still think it is important to design approaches to verify the semantic correctness of the generated model (e.g., whether each transition is physically grounded in the real world).

\section{Related Works}

\textbf{LLMs for Mechanistic Modeling.} Large language models (LLMs) have been extensively used for scientific coding tasks~\cite{tang2024biocoder, white2023assessment, zhang2025large, kwok2024utilizing}. However, most existing approaches focus on one-shot code generation, rather than iterative optimization of model structures. Recently, AlphaEvolve~\cite{novikov2025alphaevolve} proposed an evolutionary coding framework that automatically generates, evaluates, and refines computer programs for scientific and algorithmic discovery. Nevertheless, it is designed as a general-purpose system, whereas our work focuses specifically on mechanistic modeling, a subfield of scientific coding that places stronger emphasis on validity and knowledge grounding. EpiAgent~\cite{datta2026agentic} proposed an agentic framework for epidemic modeling, where the problem settings differ from ours.

The most closely related work to ours is HDTwinGen~\cite{holt2024automatically}. However, their problem setting only considers the simple hybrid model, which is a special case of the neural-integrated mechanistic modeling we consider in the NIMM benchmark.

% \textbf{LLMs for Optimization.} LLMs have been used in optimization tasks, but either in discrete prompt optimization~\cite{yang2023large}, combinatorial optimization~\cite{ye2024reevo, iklassov2024selfguiding, jiang2024llmopt, jiang2025large}, or preference optimization~\cite{lu2024discovering}. But none of them have explored 'learning from data', especially with the time series data. The most similar paper is~\cite{holt2024automatically}.

\textbf{Sequential Deep Neural Networks.} Sequential Deep Neural Networks (DNNs) can be adapted to the mechanistic modeling task. Specifically, rather than directly learning a single function $f^{mech}$, the sequential DNNs aim to learn a black-box function $f^{}$ that aims to directly predict the system state at the next time step based on a window of historical states. Typical architectures such as RNNs~\cite{elman1990finding}, LSTMs~\cite{hochreiter1997long}, and Transformers~\cite{vaswani2017attention} have been widely applied in this setting. Although sequential DNNs show promising performance in capturing complex dynamics in the data, they are heavily relying on training data for generalization and lack substantive interpretability.

\section{Conclusion}
We first propose NIMM, a benchmark with the novel neural-integrated mechanistic modeling problem setting. Based on the evaluation results on the baselines, we identify some key challenges in the current baseline methods. Motivated by these findings, we design a tree-guided agentic framework, NIMMGen, that explicitly improves search stability and solution quality during iterative refinement. Experiments across three domains demonstrate the effectiveness of the NIMMGen. We believe the paper can contribute to the community by providing a practical benchmark and an agentic system in neural-integrated mechanistic modeling. We hope the paper can serve as a call for future research in developing better methods in this direction.

\clearpage

\bibliography{references}
\bibliographystyle{plain}

\clearpage
\appendix
\section{NIMM benchmark (Public Health)}
\subsection{Overview}
The NIMM benchmark consists of four epidemiological datasets~\cite{cui2025identifying, guan2025framework, chopra2022differentiable, datta2025calypso} spanning multiple diseases, spatial resolutions, and temporal granularities. It is designed to reflect realistic challenges encountered in public health surveillance, including spatial heterogeneity and varying observation frequencies. Table~\ref{tab:dataset} summarizes the datasets used to construct the NIMM benchmark, and Figure~\ref{fig:dataset} visualizes the geolocations and aggregated time series for each dataset.

% \begin{figure*}
%     \centering
%     \includegraphics[width=0.7\linewidth]{Figures/dataset.pdf}
%     \caption{The datasets of the NIMM benchmark are collected from multi sources and multi levels.}
%     \label{fig:dataset}
% \end{figure*}
\begin{table*}[!t]
    \centering
    \caption{Statistical Description of the datasets used in the NIMM benchmark (Public Health).}
    \label{tab:dataset}
    \resizebox{\textwidth}{!}{\begin{tabular}{c|ccccccccc}
    \toprule
        Dataset Name & Data Source & Observation Period & Spatial Granularity & \# Locations & Temporal Granularity & \# Timestamps & Disease & \# Features & Target\\
    \midrule
         COVID-Bogota & Colombia NIH & 2020-04-12 - 2021-01-30 & City-level & 1 & Daily & 343 (d)  & COVID  & 11 & Infection Counts\\
         COVID-Medellin & Colombia NIH & 2020-04-12 - 2021-01-30 & City-level & 1 & Daily & 343 (d) & COVID & 11 & Infection Counts\\
         % COVID-USA & CDC & 2020-01-01 - 2021-08-21 & State-level & 51 & Daily & 602 (d) & COVID & 5 & Mortality\\
         Influenza-USA & CDC & 2017-W34 - 2021-W15\footnote{This follows the CDC/MMWR epidemiological week} & State-level & 51 & Weekly & 191 (w) & Influenza & 15 & Infection Counts\\
         MRSA-Virginia & Virginia APCD & 2016-W01 - 2021-W04 & Facility-level & 644 & Weekly & 244 (w) & MRSA  & 20 &  Infection Counts \\
         \bottomrule
    \end{tabular}}
\end{table*}

\subsection{Dataset}\label{appendix:dataset}

% \anil{not clear about "benchmark" and "assembly" doesnt seem right usage}
The benchmark covers three infectious diseases, i.e., COVID, Influenza, and Methicillin-resistant \textit{Staphylococcus aureus} (MRSA), and includes data from both macro-level surveillance systems and facility-level administrative records. The shape of each dataset can be described using three dimensions: (\# Locations, \# Timestamps, \# Features).

\paragraph{COVID-Bogota \& COVID-Medellin} These are City-level COVID-19 datasets collected from the Colombia NIH\footnote{\url{https://www.ins.gov.co/Noticias/Paginas/coronavirus-casos.aspx}}. The dataset contains daily observations of 343 days. Following~\cite{guan2025framework}, we adopt 11 features collected from heterogeneous sources, including:
\begin{itemize}[leftmargin=*]
    \item \textbf{Mobility signals}. These signals originate from records of visits to points of interest (POIs) across different regions. According to Google Mobility\footnote{\url{https://www.gstatic.com/covid19/mobility/Global_Mobility_Report.csv}}, daily changes in visits to various POI categories are collected and reported relative to the baseline period from January 3 to February 6, 2020. We extract six time-series features capturing mobility changes under different scenarios, including \texttt{retail\_and\_recreation} and \texttt{grocery\_and\_pharmacy}.
    
    \item \textbf{Google Trends signals}. Google provides search trend statistics for user-specified keywords\footnote{\url{https://trends.google.com/trends/explore?date=today}}. We collect search trends for four COVID-19–related keywords, such as ``covid-19 en Colombia,'' ``covid-19 hoy,'' and ``covid-19 Bogot\'a,'' resulting in three time-series features.
    
    \item \textbf{Infection-related context}. The Colombian National Institute of Health (NIH) reports daily COVID-19 infection and mortality counts, serving as the official data source in Colombia. From this source, we derive two time-series features.
\end{itemize}

COVID-Bogota and COVID-Medellin are both collected on the city-level. Therefore, the final shape of the two datasets is (1, 343, 11).

% \paragraph{COVID-USA} State-level COVID datasets. Following~\cite{chopra2022differentiable}, we collect signals from the following sources.

% \begin{itemize}[leftmargin=*]
%     \item \textbf{Mobility signals}. Similar to the mobility signals used in the COVID-Bogota and Covid-Medellin
% \end{itemize}

% COVID-USA is collected from 51 states in the USA. Therefore, the final shape of the Influenza-USA dataset is (51, 602, 5).

\paragraph{Influenza-USA} State-level influenza datasets. Following~\cite{chopra2022differentiable}, we adopt 14 signals.

\begin{itemize}[leftmargin=*]
    \item \textbf{Symptoms}. This contains 13 different time series symptoms features collected from the Google symptoms dataset\footnote{\url{https://github.com/GoogleCloudPlatform/covid-19-open-data/blob/main/docs/table-search-trends.md}}, including Fever, Cough, and Sore throat.
    \item \textbf{influenza-like-illness (ILI)}. This is collected by the CDC and records the influenza-like illness across different timestamps.
\end{itemize}

Influenza-USA is collected from 51 states in the USA. Therefore, the final shape of the Influenza-USA dataset is (51, 191, 14).

\paragraph{MRSA-Virginia} Facility-level MRSA datasets are collected from the Virginia All Payers Insurance Claims dataset (APCD). Built of top of~\cite{datta2025calypso}, we adopt four signals:
\begin{itemize}[leftmargin=*]
    \item \textbf{Demographical signals}. This includes the number of MRSA cases for each age group. (We have 15 age groups and therefore we have 15 features)
    \item \textbf{Prescriptions}. Number of Prescription claims listed in the insurance data, along with the timestamps.
    \item \textbf{MRSA Infections}. The number of infection counts corresponding to the MRSA.
\end{itemize}

MRSA-Virginia is collected from 644 healthcare facilities in the state of Virginia. Therefore, the final shape of the MRSA-Virginia dataset is (644, 244, 20).

\begin{figure}
    \centering
    \includegraphics[width=\linewidth]{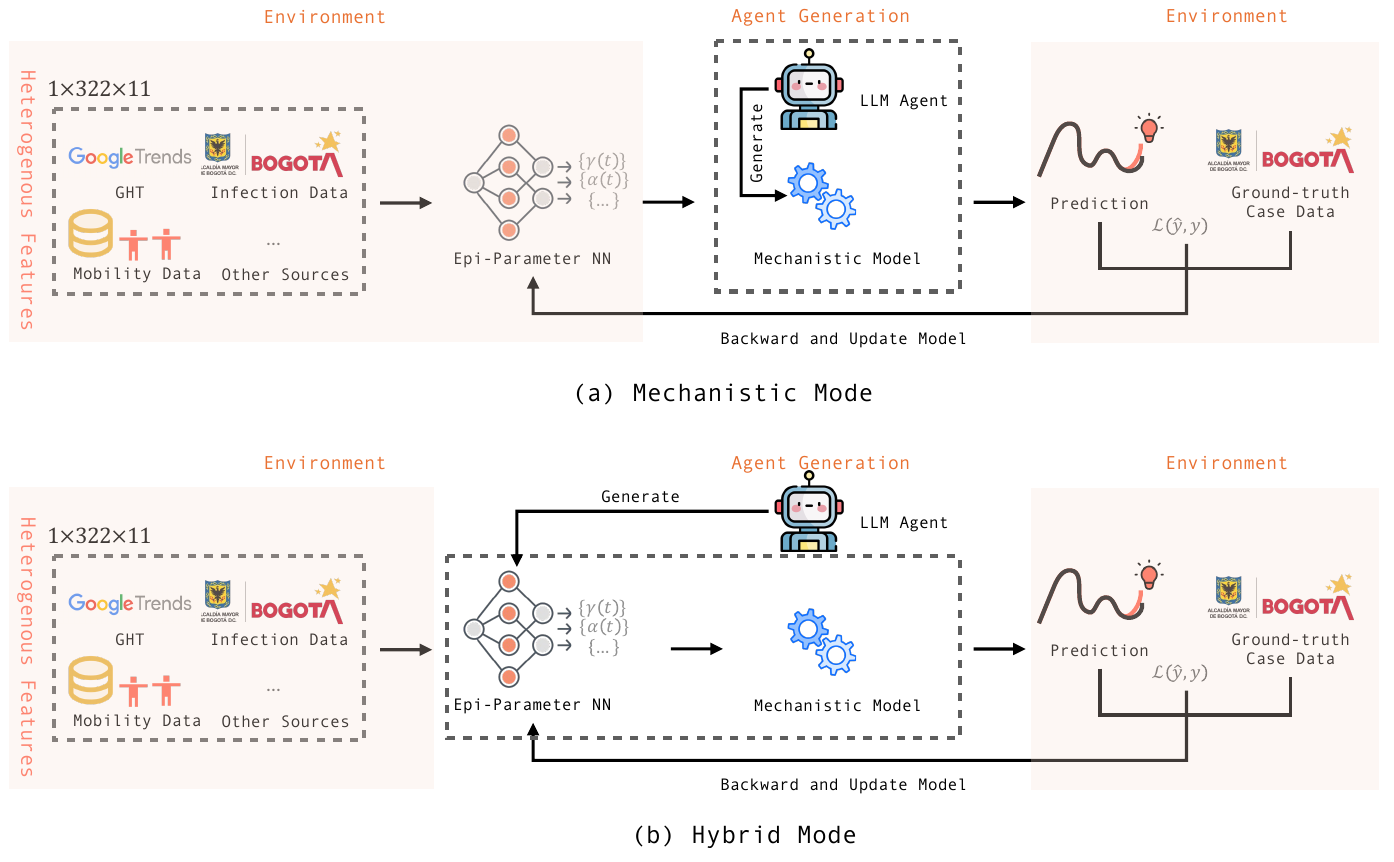}
    \caption{Experimental setup. (a) Mechanistic mode. (b) Hybrid Mode.}
    \label{fig:benchmark_setup}
\end{figure}
\subsection{Code Generation Modes} We consider two types of code generations modes: \textit{mechanistic model} and \textit{hybrid mode}. In the mechanistic mode, the agent is responsible for generating the codes for $f^{mech}$ only, while the $f^{NN}$ is predefined in the environment. In particular, in the mechanistic mode, we use the CalibNN designed by~\cite{chopra2022differentiable}, which has been demonstrated to be effective in extracting the information from heterogeneous data sources across multiple datasets. In the full model mode, the agent is responsible for generating the full model $f^{mech}(f^{NN}(\cdot))$.

\subsection{Task \& Evaluation}\label{appendix:task}
\paragraph{Task Description}
In the public health subset, we mainly focus on forecasting, which is one of the most popular tasks in epidemics. Formally, the goal in the forecasting problem is to use data till time $T$, namely $\{\mathbf{x}_t\}_{t=1}^T$,
and predict $\{\mathbf{x}_t\}_{T+1}^{T+H}$ i.e.,
the number of infections and deaths during the time period $[T+1, T+H]$, where $H$ is the forecast horizon.

\paragraph{Training and Inference} 
During training, we aim to calibrate the composite model
$f(\cdot) = f^{mech}({f_{\beta}^{NN}(\cdot)})$
from the training data $\{D_{:,t}\}_{t=1}^T$. Specifically, the neural network
\( \theta = f_{\beta}^{NN}(\{D_{:,t}\}_{t=1}^T) \)
takes the entire training sequence as input and predicts the parameters of the mechanistic model.
These predicted parameters \( \theta \) are then used to instantiate the mechanistic model \( f^{mech} \),
which is executed for \( T \) time steps to generate simulations for the target state (e.g., infection counts), i.e., $\{\hat{y}_{:, t}\}_{t=1}^T$. The training objective is to
minimize the root mean squared error (RMSE) between the simulated target values
\( \{\hat{y}_{:,t}\}_{t=1}^T \)
and the ground-truth target observations 
\( \{y_{:,t}\}_{t=1}^T \):
\begin{equation}
\begin{split}
        \beta^* = & \arg \min_{\beta} \text{RMSE}(\{\hat{y}_{:, t}\}_{t=1}^T, \{y_{:, t}\}_{t=1}^T), \\
        \text{where} \quad & \{\hat{y}_{:, t}\}_{t=1}^T 
        = f^{mech}({f_{\beta}^{NN}(\{D_{:,t}\}_{t=1}^T)}) .
\end{split}
    \label{eqn:train}
\end{equation}
At inference time, we perform forecasting by running the calibrated composite model for \( T+H \) steps.
The first \( T \) steps reproduce the observed period, while the additional \( H \) steps generate predictions
for the future horizon \( [T+1, \ldots, T+H] \).

\paragraph{Real-time Evaluation}
\begin{figure}
    \centering
    \includegraphics[width=0.6\linewidth]{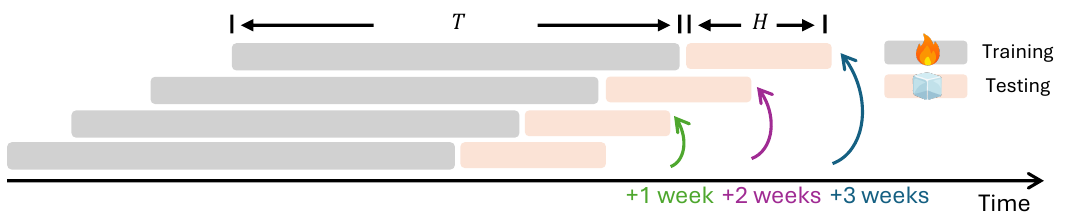}
    \caption{Real-time evaluation Setup.}
    \label{fig:real_time}
\end{figure}
Following the popular real-time evaluation pipelines in the epidemics~\cite{chopra2022differentiable, adhikari2019epideep, kamarthi2022camul, shaman2013real}, we mainly evaluate the performance of the constructed models using the \textit{real-time forecasting} setup. We simulate real-time forecasting by forcing the models to train only using data available until each of the prediction dates and make predictions for 4 weeks ahead in the future. Data revisions in public health data are large and may affect evaluation and conclusions~\cite{cramer2022evaluation, kamarthi2021back2future}, therefore, we utilize fully revised data following previous papers on methodological advances.

Specifically, we shift the training time window to 1, 2, and 3 weeks ahead to simulate the real-time setting. Figure~\ref{fig:real_time} visualizes the real-time evaluation setups.

\paragraph{Environment Setup}
The learning rate of the training is set as $5e^{-4}$ with AdamW as the optimizer. The training iteration is set to 1000 to ensure convergence. The $H$ is set to be 8 (weeks) for Influenza-USA and MRSA-Virginia datasets, and 28 (days) for the COVID-Bogota and COVID-Medellin datasets. The $T$ is chosen depending on different datasets. For the COVID-Bogota and COVID-Medellin, the default $T$ is 294 days (42 weeks); For the Influenza-USA, the default $T$ is 191 (weeks); For the MRSA-Virginia, the default $T$ is 244 (weeks). 

To use the COVID-Bogota and COVID-Bogota dataset for real-time evaluation, the training period will shift from [1,...,294] to [21,...,315], and the testing period will accordingly shift from [295,...,322] to [316,...,343].

To use the Influenza-USA dataset for real-time evaluation, the training period will shift from [1,...,179] to [4,...,182], and the testing period will accordingly shift from [180,...,188] to [183,...,191].

To use the MRSA-Virginia dataset for real-time evaluation, the training period will shift from [1,...,232] to [4,...,235], and the testing period will accordingly shift from [233,...,241] to [236,...,244].

\paragraph{Experimental Setup}
The default iteration number for the NIMMGen pipeline is set to 40. All experiments are conducted on two A100 GPUs. We use GPT-4.1 as the default backbone LLM.

\subsection{Baselines}
As stated in the main text, we opt to use the following three types of baselines.

\subsubsection{Neural Sequence Models}
\paragraph{LSTM Model} We adopt a simple LSTM model consisting of one LSTM layer with hidden size 32, followed by a dense layer with hidden size 16, a rectified linear unit (ReLU) activation function, and a dropout layer with a dropout rate of 0.2. The output layer is a dense layer with linear activation and $L_2$ kernel regularization with a penalty factor of 0.01. The input dataset $\{D_t\}_{t=1}^T$ is first segmented into overlapping five-timestamp windows, where the first four timestamps $(D_{t-4}, \ldots, D_{t-1})$ are used as input features and the last timestamp $D_t$ is used as the prediction target. Formally, the model learns a mapping
\[
\hat{D}_t = f^{\text{LSTM}}(D_{t-4}, D_{t-3}, D_{t-2}, D_{t-1}).
\]

In the inference stage, we generate $H$-step-ahead forecasts in an auto-regressive manner. Starting from the last observed window $(D_{T-3}, D_{T-2}, D_{T-1}, D_T)$, the model predicts
\[
\hat{D}_{T+1} = f^{\text{LSTM}}(D_{T-3}, D_{T-2}, D_{T-1}, D_T).
\]
This prediction is then fed back into the input window to generate subsequent forecasts:
\begin{equation}
\begin{split}
    \hat{D}_{T+h} = f^{\text{LSTM}}(\hat{D}_{T+h-4}, \hat{D}_{T+h-3}, \hat{D}_{T+h-2}, \hat{D}_{T+h-1})&, \\
    h = 2, \ldots, H &,
\end{split}
\end{equation}
where observed values are used when available and predicted values are used once the forecasting horizon extends beyond the training data. This iterative process yields the final $H$-step forecast $\{\hat{D}_{T+1}, \ldots, \hat{D}_{T+H}\}$.

\paragraph{Transformer Model} We first encode the input dimension of the observations into an embedding vector dimension of size 64 through a linear network, followed by the addition of a standard positional encoder. Then, this is fed into the Transformer encoder layers, which had a head size of 4 and contain three layers of Transformer blocks. The final output is mapped back to the input dimension by using an additional linear layer. The training and the inference strategies are similar to those in the LSTM model.

\subsubsection{Existing Neural-integrated Mechanistic Models}

\paragraph{Grad Meta-population Model}

% \anil{this is not a pure mechanistic model. It is hybrid where we fix the mechanistic model structure and require it to just learn the parameters.
% }
A meta-population model requires two key inputs: a contact matrix $\mathcal{C} \in \mathbb{R}^{m \times m}$ and a population vector $N \in \mathbb{R}^{m}$, where $m=|\mathcal{G}|$ is the number of groups. 
Each entry $C_{i,j} \in \mathcal{C}$ represents the contact probability between the populations in groups $i$ and $j$, while each entry $N^j \in N$ records the population of group $j$. Following~\cite{venkatramanan2019optimizing}, the effective number of infected individuals and the effective population of group $j$ after daily movement are defined as:
\begin{equation}
    I^{\textsc{eff}}_j = \sum_{i} \mathcal{C}_{i,j} I_i, \quad N^{\textsc{eff}}_j = \sum_i \mathcal{C}_{i,j} N_i
\end{equation}
% where $I^{\textsc{eff}}\in\mathbb{R}^m$ represents the effective number of infections in each group, 
where "\textsc{eff}" (effective) refers to infections accounted for, including infected individuals between different subpopulations due to transmissibility and mobility. The conditional force of infection for group $j$ is then given by:
\begin{equation}
    \beta^{\textsc{eff}}_j = \beta \frac{I^{\textsc{eff}}_j}{N^{\textsc{eff}}_j}.
\end{equation}
For each group $i \in [1,...,m]$, the epidemic evolution $\mathcal{Z}_i = [S_i, E_i, \allowbreak I_i, R_i, M_i]$ from day $t$ to day $t+1$ is formulated as:
\begin{equation}
\begin{aligned}
    \mathcal{Z}_i(t+1) &= \mathcal{Z}_i(t) + \Delta {\mathcal{Z}_i(t)}, \quad where \\
    % \begin{split}
    \Delta {S_i(t)} &= -\sum_{j=1}^{m} C_{i,j} \beta^{\textsc{eff}}_j S_i(t) + \delta(t) R_i(t),\\
    \Delta {E_i(t)} &= \sum_{j=1}^{m} C_{i,j} \beta^{\textsc{eff}}_j S_i(t) - \alpha(t) E_i,\\
    \Delta {I_i(t)} &= \alpha(t) E_i(t) - (\gamma(t) + \eta(t)) I_i(t), \\
    \Delta {R_i(t)} &= \gamma(t) I_i(t) + (1-\delta(t)) R_i(t), \\
    \Delta {M_i(t)} &= \eta(t) I_i(t). 
    % \end{split}
    \label{eqn:transmission}
\end{aligned}
\end{equation}
where our forecasting objective $\Delta I_i(t)$ represents the number of newly infected individuals in age group $i$ at time $t$. Running the meta-population model for $T$ time steps, it outputs the aggregated new cases over all the age groups for the entire training period:
\begin{equation}
    \hat{y}_t := \sum_{i=1}^m \Delta I_i(t), \forall 1 \leq t \leq T.
\end{equation}

Following~\cite{guan2025framework}, we calibrate the model with a CalibNN. We follow the same implementations as outlined in GitHub\footnote{\url{https://github.com/GuanZihan/GradMetapopulation.git}}.

\subsubsection{LLM-based Generations} 
\paragraph{Zero-shot} For zero-shot generation, we prompt the LLM model with the exact task description and the skeleton codes as our pipeline. However, it only generates the code once and incorporates no criticism from the reflection agent.

\paragraph{HDTwinGen~\cite{holt2024automatically}} We adopt the same implementation as described in the official GitHub~\footnote{\url{https://github.com/samholt/HDTwinGen}}.
% \anil{and this is for addition of the functions, right?}

\section{NIMM (Clinical Health) Setup}\label{appendix:cancer}

\subsection{Dataset}
Following~\cite{holt2024automatically}, we include a biomedical Pharmacokinetic-Pharmacodynamic (PKPD) benchmark derived from a state-of-the-art mechanistic model of lung cancer tumor growth~\cite{geng2017prediction}, which has been widely adopted in prior studies on treatment effect modeling and counterfactual simulation~\cite{bica2020estimating, seedat2022continuous, melnychuk2022causal}. The benchmark simulates tumor progression under different treatment strategies, including: (1) no treatment, (2) chemotherapy only, and (3) combined chemotherapy and radiotherapy. We use this environment to evaluate whether neural-integrated mechanistic models can correctly recover clinically meaningful treatment dynamics and generate stable long-term forecasts.

The underlying system models the temporal evolution of tumor volume $x(t)$ after diagnosis using a mechanistic differential equation that combines intrinsic tumor growth dynamics with treatment from chemotherapy and radiotherapy. The model has two binary treatments: (1) radiotherapy $u_t^r$ and (2) chemotherapy $u_t^c$.

\begin{equation}
\frac{dx(t)}{dt}
=
\left(
\rho \log\left(\frac{K}{x(t)}\right)
-
\beta_c c(t)
-
\left(\alpha_r d(t) + \beta_r d(t)^2\right)
\right)x(t),
\label{eqn:pkpd}
\end{equation}

where $K=30$ denotes the tumor carrying capacity, $\rho=7.00\times10^{-5}$ controls the intrinsic growth rate, $\alpha_r=0.0398$ denotes the linear radiotherapy cell-kill parameter, $\beta_r$ denotes the quadratic radiotherapy cell-kill parameter and is set such that $\alpha_r/\beta_r=10$, $\beta_c=0.028$ denotes the chemotherapy cell-kill parameter, $c(t)$ represents the chemotherapy drug concentration, and $d(t)$ denotes the radiotherapy dosage.

The chemotherapy concentration follows an exponential decay process with a one-day half-life:

\begin{equation}
\frac{dc(t)}{dt} = -0.5c(t).
\end{equation}

The chemotherapy action is modeled as a binary intervention that increases the chemotherapy concentration $c(t)$ by $5.0\,\text{mg}/\text{m}^3$ of Vinblastine when administered. The radiotherapy action corresponds to delivering a $2.0\,\text{Gy}$ fraction of ionizing radiation at timestep $t$, where Gy denotes the Gray radiation dose unit.

The treatment assignment process additionally incorporates time-varying confounding. Specifically, chemotherapy and radiotherapy administrations are sampled as Bernoulli random variables whose probabilities depend on the current tumor diameter. The associated probabilities, $p_c(t)$ and $p_r(t)$, represent the probability for the two actions:

\begin{equation}
p_c(t)=\sigma\left(\frac{\gamma_c}{D_{\max}}(\bar{D}(t)-\delta_c)\right),\quad p_r(t)=\sigma\left(\frac{\gamma_r}{D_{\max}}(\bar{D}(t)-\delta_r)\right),
\label{eqn:action_policy}
\end{equation}

where $D_{\max}=13\,\text{cm}$ denotes the largest tumor diameter, $\delta_c=\delta_r=D_{\max}/2$, $\bar{D}(t)$ denotes the mean tumor diameter, and $\gamma_c=\gamma_r=2$ controls the extent of time-varying confounding. Following the original simulation protocol, we generate patient trajectories using Euler-based forward simulation over 60 days. Each trajectory contains sequential observations of tumor progression together with treatment interventions. The benchmark is particularly challenging for mechanistic modeling because the observed outcomes emerge from nonlinear interactions between intrinsic tumor growth, pharmacokinetic decay, radiotherapy response, and time-dependent treatment allocation policies. Consequently, successful modeling requires both accurate dynamical forecasting and robust handling of treatment-dependent temporal confounding.

Using the above Cancer PKPD model, we sample $N=1000$ trajectories where we sample their initial tumor volumes from a uniform distribution $x(0) \sim \mathcal{U}(0, 1149)$ and use the Cancer PKPD Equation~\ref{eqn:pkpd} along with the action policy in Equation~\ref{eqn:action_policy} to forward simulate patient trajectories for 60 days with an Euler stepwise solver. At each timestamp, the treatment action $u_t = (u_t^c, u_t^r)$ consists of the chemotherapy and radiotherapy interventions sampled from the treatment assignment policy.

This results in a trajectory dataset
 
\begin{equation}
\mathcal{D}
=
\left\{
\tau^{(i)}
\right\}_{i=1}^{N},
\quad
\tau^{(i)}
=
\left\{
(x_t^{(i)}, u_t^{(i)}, x_{t+1}^{(i)})
\right\}_{t=0}^{T-1}.
\end{equation} 

Following the prior work, we independently repeat the above process three times for generating training, validation, and testing datasets, denoted as $D_{train}$, $D_{val}$, and $D_{test}$.

\subsection{Task Formulation}
Given the current tumor state $x_t \in \mathbb{R}^{d_\mathcal{X}}$ and treatment actions $u_t$ , the objective is to train a neural-integrated mechanistic model $f(\cdot) = f^{mech}(f^{NN}_{\beta}(\cdot))$ to predict the next state $x_{t+1}$. 

Formally, we model the system dynamics as a composite neural-integrated mechanism:

\begin{equation}
\hat{x}_{t+1} = f^{\text{mech}} \big(x_t, u_t; f^{\text{NN}}_{\beta}(x_t, u_t)\big),
\end{equation}

where the neural network component $f^{\text{NN}}_{\beta}(x_t, u_t)$ is used to predict the parameters for the mechanistic component $f^{mech}$, and the mechanistic component then evolves the tumor dynamics according to the inferred parameters and treatment conditions.

The model is trained by minimizing the loss function given below:

\begin{equation}
\mathcal{L}
=
\frac{1}{N}
\sum_{i=1}^{N}
\sum_{t=0}^{T-1}
\left\|
\hat{x}_{t+1}^{(i)}
-
x_{t+1}^{(i)}
\right\|_2^2,
\end{equation}

where $N$ denotes the number of samples in the training dataset, the upper script $i$ denote the sample index $i \leq N$, $x_{t+1}^{i}$ denotes the ground-truth value, and $\hat{x}^{i}_{t+1}$ denotes the predicted value by the neural-integrated model $f(\cdot)$.

\subsection{Code Generation Modes} As in the public health subset, we also consider two types of code generation modes: \textit{mechanistic model} and \textit{hybrid mode}. In the mechanistic mode, the agent is responsible for generating the codes for $f^{mech}$ only, while the $f^{NN}$ is predefined in the environment. In particular, in the mechanistic mode, we use a simple 2-layer MLP as $f^{NN}$. In the full model mode, the agent is responsible for generating the full model $f^{mech}(f^{NN}(\cdot))$.

% \anil{would be useful to say what the mechanistic model is.
% How are these being searched on the tree?}

\section{NIMM (Materials Science) Setup}\label{appendix:materials}

\subsection{Task Formulation}
The considered task is to predict the temperature-dependent yield strength of single-phase Face-Centered Cubic (FCC) / Body-Centered Cubic (BCC) high-entropy alloys (HEAs) by integrating a neural network with a physics-based mechanistic model~\cite{yin2019first, maresca2020mechanistic}. Similar to the setup in NIMM, the neural network component is used to predict key mechanistic parameters from alloy composition data $D$. Then these predicted parameters are used to propagate through a mechanistic model, yielding the final yield strength prediction.

The Materials subset follows the setup of~\cite{liu2022integrating}, where the task is to train a neural-integrated mechanistic model $f(\cdot) = f^{mech}(f^{NN}_{\beta}(\cdot))$ that predicts alloy yield strength $\hat{y}_i$ for the given descriptor features $x_i$ of the alloy $i$. The descriptors include features such as valence electron concentration and atomic volume. Following the neural-integrated mechanistic modeling framework, the prediction $\hat{y}_i$ is given by:

\begin{equation}
    \hat{y}_i = f^{mech}(f^{NN}_{\beta}(x_i)),
\end{equation}

where $f^{NN}_{\beta}$ is parameterized by $\beta$, and $f^{mech}$ is a physics-inspired mechanistic model that computes the alloy yield strength.

% \anil{are new models being considered here?}

\subsection{Datasets}
For FCC alloys, the dataset $D$ contains 29 HEAs and room temperature yield strength, where each input feature characterizes composition-based descriptors derived from elemental properties~\cite{ward2016general, liu2022integrating} (e.g., valence electrons, atomic volume). For BCC alloys, the dataset $D$ contains 130 HEAs with temperature-dependent yield strength data, where each input feature also characterizes composition-based descriptors derived from elemental properties~\cite{borg2020expanded} (e.g., valence electrons, atomic volume).

\subsection{Code Generation Modes} We also consider two types of code generation modes: \textit{mechanistic model} and \textit{hybrid mode}. In the mechanistic mode, the agent is responsible for generating the codes for $f^{mech}$ only, while the $f^{NN}$ is predefined in the environment. In particular, in the mechanistic mode, we use a simple 2-layer MLP as $f^{NN}$. In the full model mode, the agent is responsible for generating the full model $f^{mech}(f^{NN}(\cdot))$.

\section{Algorithm} \label{appendix:algorithm}
\paragraph{Initialization from Skeleton Code.}
  NIMMGen takes as input a skeleton code template $S^0$, which specifies only the function signatures and structural interface of the target model, without providing a concrete implementation. Before tree expansion
  begins, an initialization LLM generates a simple executable model $f^0$ conditioned on the system prompt $C$ and the skeleton $S^0$. This initialization step is constrained to produce a minimal and non-complex
  implementation, so that the search starts from a valid but lightweight baseline model. The resulting model $f^0$ is used as the root node of the search tree.
  
\paragraph{Tree-Guided Search.}
  Algorithm~\ref{alg:nimmgen_outer} summarizes the outer-loop optimization procedure of NIMMGen. Starting from a single initialized skeleton model $f^0$, the framework builds a search tree by iteratively expanding one
  parent node at a time. At iteration $g$, NIMMGen first constructs the valid and buggy frontiers under depth, child-capacity, and validation-quality constraints. It then selects an expandable parent node according to
  the tree policy: valid nodes are prioritized by predictive performance, while buggy nodes may be revisited during designated debug rounds or when no valid node remains. After a new child model is generated and
  evaluated, the tree and memory are updated, and the current best valid model $f^\star$ is refreshed accordingly.

\paragraph{Expansion and Evaluation.}
  Algorithm~\ref{alg:nimmgen_inner} describes the inner-loop expansion step for a selected parent node. Given the system prompt $C$, the selected node $v_p^g$, and the accumulated tree memory, NIMMGen first assembles a
  node-specific context $\mathcal{M}^g$ containing the relevant historical information. A plan agent then proposes an atomic refinement action $E^g$, such as adding a compartment, revising a dependency, or modifying a
  mechanistic term. Conditioned on $C$ and $\mathcal{M}^g$, the modeling agent generates a new neural-integrated mechanistic model $f^g = f^{\mathrm{mech}}(f_\beta^{\mathrm{NN}})$. The generated model is
  subsequently trained and evaluated in the benchmark environment, producing its objective value $J(f^g)$, validation loss $L_{\mathrm{val}}(f^g)$, and bug indicator $\mathrm{bug}(f^g)$, which are returned to the outer
  search loop.
  
\begin{algorithm}[t]
\small
  \caption{Tree-Guided Search in NIMMGen}
  \label{alg:nimmgen_outer}
  \begin{algorithmic}[1]
  \Require system prompt $C$, skeleton code $S$, iteration budget $G$, maximum depth $D_{\max}$, maximum children per node $C_{\max}$, validation threshold $\tau_{\mathrm{val}}$, debug period $K$
  \Ensure best model $f^\star$

  \State Generate an initial simple implementation
  \[
  f^{0} \sim \mathrm{LLM}_{\mathrm{init}}\!\left(C, S\right)
  \]
  \State $\mathcal{T}^{0} \gets \{f^{0}\}$
  \State Evaluate $f^{0}$ and initialize memory $\mathcal{M}^{0}$
  \State $\mathcal{T}^{0}_{\mathrm{valid}} \gets \{f \in \mathcal{T}^{0} : \mathrm{bug}(f)=0\}$; $\mathcal{T}^{0}_{\mathrm{buggy}} \gets \{f \in \mathcal{T}^{0} : \mathrm{bug}(f)=1\}$
  \State $f^\star \gets \arg\min_{f \in \mathcal{T}^{0}_{\mathrm{valid}}} J(f)$

  \For{$g=1,\dots,G$}
      \State
      \[
      \mathcal{F}^{g}_{\mathrm{valid}}
      =
      \left\{
      v \in \mathcal{T}^{g-1}_{\mathrm{valid}}
      :
      d(v)<D_{\max},\;
      c(v)<C_{\max},\;
      L_{\mathrm{val}}(v)\le \tau_{\mathrm{val}}
      \right\}
      \]
      \State
      \[
      \mathcal{F}^{g}_{\mathrm{buggy}}
      =
      \left\{
      v \in \mathcal{T}^{g-1}_{\mathrm{buggy}}
      :
      d(v)<D_{\max},\;
      c(v)<C_{\max}
      \right\}
      \]
      \If{$\mathcal{F}^{g}_{\mathrm{valid}}\neq\emptyset$ and $g \not\equiv 0 \pmod K$}
          \[
          v_p^g
          =
          \arg\min_{v\in \mathcal{F}^{g}_{\mathrm{valid}}}
          \bigl(J(v),-d(v),c(v),\mathrm{id}(v)\bigr)
          \]
      \Else
          \[
          v_p^g
          =
          \arg\min_{v\in \mathcal{F}^{g}_{\mathrm{buggy}}}
          \bigl(d(v),c(v),\mathrm{id}(v)\bigr)
          \]
      \EndIf

      \State $(f^{g},J(f^{g}),L_{\mathrm{val}}(f^{g}),\mathrm{bug}(f^{g})) \gets
      \mathrm{ExpandAndEvaluate}(C,v_p^g,\mathcal{T}^{g-1},\mathcal{M}^{g-1})$

      \State $\mathcal{T}^{g}\gets \mathcal{T}^{g-1}\cup\{f^{g}\}$
      \State $\mathcal{M}^{g}\gets \Psi(\mathcal{M}^{g-1},f^{g})$
      \State $\mathcal{T}^{g}_{\mathrm{valid}} \gets \{f \in \mathcal{T}^{g} : \mathrm{bug}(f)=0\}$
      \State $\mathcal{T}^{g}_{\mathrm{buggy}} \gets \{f \in \mathcal{T}^{g} : \mathrm{bug}(f)=1\}$

      \If{$\mathrm{bug}(f^{g})=0$ and $L_{\mathrm{val}}(f^{g}) < L_{\mathrm{val}}(f^\star)$}
          \State $f^\star \gets f^{g}$
      \EndIf
  \EndFor

  \State \Return $f^\star$
  \end{algorithmic}
\end{algorithm}

\begin{algorithm}[t]
\small
  \caption{$\mathrm{ExpandAndEvaluate}(C,v_p^g,\mathcal{T}^{g-1},\mathcal{M}^{g-1})$}
  \label{alg:nimmgen_inner}
  \begin{algorithmic}[1]
  \Require system prompt $C$, selected parent node $v_p^g$, tree $\mathcal{T}^{g-1}$, memory $\mathcal{M}^{g-1}$
  \Ensure child model $f^g$, objective $J(f^g)$, validation loss $L_{\mathrm{val}}(f^g)$, bug flag $\mathrm{bug}(f^g)$

  \State Assemble context
  \[
  \mathcal{X}^{g}
  =
  \Phi\!\left(
  C,\;
  \mathcal{S}(v_p^g),\;
  \mathcal{M}^{g-1}
  \right)
  \]
  \State
  \[
  E^{g}\sim \texttt{LLM}_{\mathrm{plan}}(\mathcal{X}^{g})
  \]
  \State
  \[
  f^{g}
  =
  f^{\mathrm{mech}}\!\left(f_{\beta}^{\mathrm{NN}}\right)
  \sim
  \mathrm{LLM}_{m}\!\left(C,\mathcal{X}^{g},E^{g}\right)
  \]
  \State
  \[
  \bigl(
  J(f^{g}),
  L_{\mathrm{val}}(f^{g}),
  \mathrm{bug}(f^{g})
  \bigr)
  \gets
  \mathrm{Eval}(f^{g})
  \]
  \State \Return $\bigl(f^{g},J(f^{g}),L_{\mathrm{val}}(f^{g}),\mathrm{bug}(f^{g})\bigr)$
  \end{algorithmic}
  \end{algorithm}
\section{Detailed Setups}
% \subsection{RAG Tool}\label{appendix:rag}
% We construct the RAG database using \texttt{Qwen/Qwen3-Embedding-0.6B} as the embedding model. During the inference stage of the RAG module, the top-3 most similar code chunks are retrieved and appended to the task prompt as contextual augmentation. The code repository is built by collecting epidemic models proposed in papers published over the past five years and incorporating their released implementations, when available, to construct the RAG database.

\subsection{Open-source Models Setup} \label{appendix:open-source}
To improve the reproducibility of the NIMMGen, we set the temperature to 0 for all experiments using open-source models. The \texttt{max\_tokens} parameter is set to 2048. We use vLLM~\cite{kwon2023efficient} to deploy the open-source models and adopt the OpenAI API schema to communicate with them. All experiments are conducted on two A100 GPUs. The two open-source models are \texttt{Qwen/Qwen3-Coder-30B-A3B-Instruct}\footnote{\url{https://huggingface.co/Qwen/Qwen3-Coder-30B-A3B-Instruct}} and \texttt{Qwen/Qwen2.5-Coder-32B-Instruct}\footnote{\url{https://huggingface.co/Qwen/Qwen2.5-Coder-32B-Instruct}} respectively.

% \begin{table}[!h]
%     \centering
%     \caption{RMSE comparison between the NIMMGen and the pure black-box neural networks on the yield strength prediction task.}
%     \label{tab:materials}
%     \begin{tabular}{c|cc}
%     \toprule
%         & Neural Network & Ours \\
%         \midrule
%          FCC &  ${213.13}_{\pm 58.51}$ & $\textbf{139.19}_{\pm 29.25}$\\
%          BCC & $235.10_{\pm 15.55}$ & $\textbf{180.13}_{\pm 68.99}$\\
%          \bottomrule
%     \end{tabular}

% \end{table}

% \section{More Results}

\begin{table}[!h]
    \centering
    \caption{Performance Comparison under Different Iterations with the Influenza-USA dataset in mechanistic mode.}
    \label{tab:iteration}
    \resizebox{0.6\linewidth}{!}{\begin{tabular}{c|ccccc}
    \toprule
         & 10 & 20 & 40 & 80 \\
         \midrule
         Influenza-USA & $13.16_{\pm 3.81}$ & $11.13_{\pm 1.26}$ & $7.57_{\pm 2.05}$ & $6.25_{\pm 1.78}$\\
         \bottomrule
    \end{tabular}}
\end{table}

\begin{figure}
    \centering
    \includegraphics[width=0.4\linewidth]{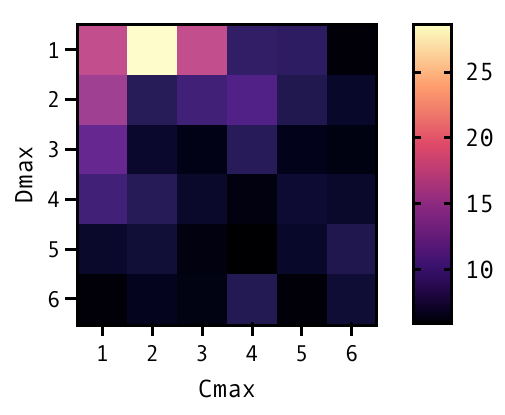}
    \caption{Performance (measured by RMSE) of NIMMGen on the Influenza-USA dataset (mechanistic mode) with different choices of $D_{max}$ and $C_{max}$. The performance becomes stably good with $C_{max} \geq 3$ and $D_{max} \geq 3$.}
    \label{fig:heatmap}
\end{figure}

\section{Prompts}\label{appendix:prompts}
\subsection{Reflection Prompt}
\begin{minted}[breaklines, fontsize=\small, frame=lines]{python}
def build_tree_reflection_prompt(
    *,
    env_name: str,
    iteration: int,
    global_memory: str,
    target_loss: Optional[float],
) -> str:
    target_text = "as low as possible" if target_loss is None else f"below {target_loss:.6f}"
    return dedent(
        f"""
        You are reviewing a mechanistic-model tree search for {env_name}.
        The objective is to drive the test loss {target_text}.

        Global tree memory:
        ```
        {global_memory}
        ```

        Call the `ReflectionSubmission` tool with short, concrete guidance for the next expansion.
        Focus on:
        - which structural edit classes are promising,
        - whether the current best branch should be deepened,
        - which buggy nodes are worth repairing,
        - and which repeated failure patterns should be avoided.

        This is reflection round {iteration}.
        """
    ).strip()
\end{minted}

% \subsection{Error Correction Prompt}
% \begin{tcolorbox}[title=Error Correction Prompt]
% Summarize code execution failures into concise bullets with key code fragments, error messages, and actionable fixes.

% Condense the repeated failures below. For each, keep a short code excerpt, the core error message, and a specific fix.
% \{errors\_text\}

% The following section records generations that failed to run in history. Reflect on the error messages also and describe how to fix the corresponding code so it executes correctly:

% \#\#\#\#\#\#\#\#\#\#\#\#

% \{joined\_errors\}

% \#\#\#\#\#\#\#\#\#\#\#\#
% \end{tcolorbox}

\subsection{Skeleton Codes Prompt}

The skeleton codes are very similar across different datasets. Here, we use Influenza-USA as an example.

\begin{minted}[breaklines, fontsize=\small, frame=lines]{python}
class StateDifferential:
    def __init__(self, population, theta):
        # This section provides prefilled code that enables the model to incorporate meta-population dynamics.
        # Initialize the parameters
        self.device = 'cuda:0'
        
        # migration matrix defines a matrix describing the contact probability among N locations. Its size is (N, N)
        self.migration_matrix = theta
        
        # num_agents defines a vector describing the population of each location. Its size is (N)
        self.num_agents = population
        
        # self.num_patches is N.
        self.num_patches = self.migration_matrix.shape[0]

    def step(self, params, T):
        # TODO: Fill in the code here
        # You can use any number of parameters from ['beta', 'kappa', 'symprob', 'epsilon', 'alpha', 'gamma', 'delta', 'mor'] to construct the mechanistic model.
        # The shape of `params`: (self.num_patches, T, 8)
        # Note that 'beta_t', 'kappa_t', 'symprob_t', 'epsilon_t', 'alpha_t', 'gamma_t', 'delta_t', and 'mor_t' are all of shape (self.num_patches)
        # **You MUST NOT using any sub loops exceept inside this loop**
        for t in range(T):
            (
                beta_t,
                alpha_t,
                gamma_t,
                delta_t,
                kappa_t,
                epsilon_t,
                symprob_t,
                mor_t,
            ) = params[:, t, :].unbind(dim=-1)
            S_t = S[-1]
            E_t = E[-1]
            I_t = I[-1]
            R_t = R[-1]
            M_t = M[-1]

            # TODO: Fill in the code here
            # - Remember this is a deterministic ODE Simulator
            # - Return `I` as tensors, where `I` denotes the simulated infected counts along different time stamps; `I` should be the shape of (`T`+1, self.num_patches); params have a grad_fn, so make sure that `I` also remains in the computation graph.
            
        return I
\end{minted}

Note that the `population' and the `theta' are two constant vectors passed by the environment. We used the ones collected from a popular source\footnote{\url{https://github.com/NSSAC/patchflow-data/tree/main}}.

\section{Qualitative Analysis on the Generated Mechanistic Codes}

\subsection{Evolving Code Examples}\label{appendix:evolve}

Table~\ref{tab:influenza_search_summary} summarizes the major structural refinements explored during a run in the Influenza-USA dataset. Rather than listing every generated node, the table highlights the best refinement discovered at each effective stage of the search, together with its validation and test losses. This provides a compact view of how the search progressed from simple compartmental baselines toward more expressive neural-integrated mechanistic models.

Several findings emerge from Table~\ref{tab:influenza_search_summary}. First, the largest early improvement came from introducing external importation into the exposed compartment, which substantially reduced the validation loss (17.03 $\rightarrow$ 8.95) and suggests that sporadic external introductions were essential for explaining the observed influenza dynamics. Second, branch-level exploration was important: although one strong intermediate branch introduced cumulative incidence tracking (row 3), the final best model (row 5) emerged from a different branch that combined patch-specific importation, explicit hospitalization, and patch-specific hospital mortality. This indicates that preserving multiple mechanistic hypotheses during search was beneficial.

\begin{table*}[t]
\centering
\caption{Evolution of Mechanistic Model Design and Validation Loss for the Influenza-USA dataset}
\label{tab:influenza_search_summary}
\resizebox{\linewidth}{!}{\begin{tabular}{p{1.5cm}p{3.2cm}p{5.8cm}cc}
  \toprule
  Depth & Parent model & Best refinement explored at this stage & Test.\ loss & Val loss \\
  \midrule
  1 & Seed models & Best initial seed: \textsc{seird} baseline with latent exposure, recovery, and mortality & 11.79 & 17.03 \\
  2 & \textsc{seird} & Add external importation into the exposed compartment $E$ to model sporadic introductions & 11.63 & 8.95 \\
  3 & \textsc{seird} + importation & Add a cumulative incidence compartment $C$ to track total new infections over time & 8.25 & 7.68 \\
  4 & \textsc{seird} + patch-specific importation & Add a hospitalization compartment $H$ to capture severe-case progression & 9.06 & 8.89 \\
  5 & \textsc{seirhd} + patch-specific importation & Make hospital mortality patch-specific to capture spatial heterogeneity in severe outcomes & 8.27 & 6.23 \\
  6 & Best depth-5 model & Further refinements explored: time-varying transmission, patch-specific hospitalization, direct out-of-hospital mortality, and patch-specific waning immunity; none improved over depth 5 & 7.97 & 7.54 \\
  \bottomrule
  \end{tabular}}
\end{table*}

\begin{figure}
    \centering
    \includegraphics[width=\linewidth]{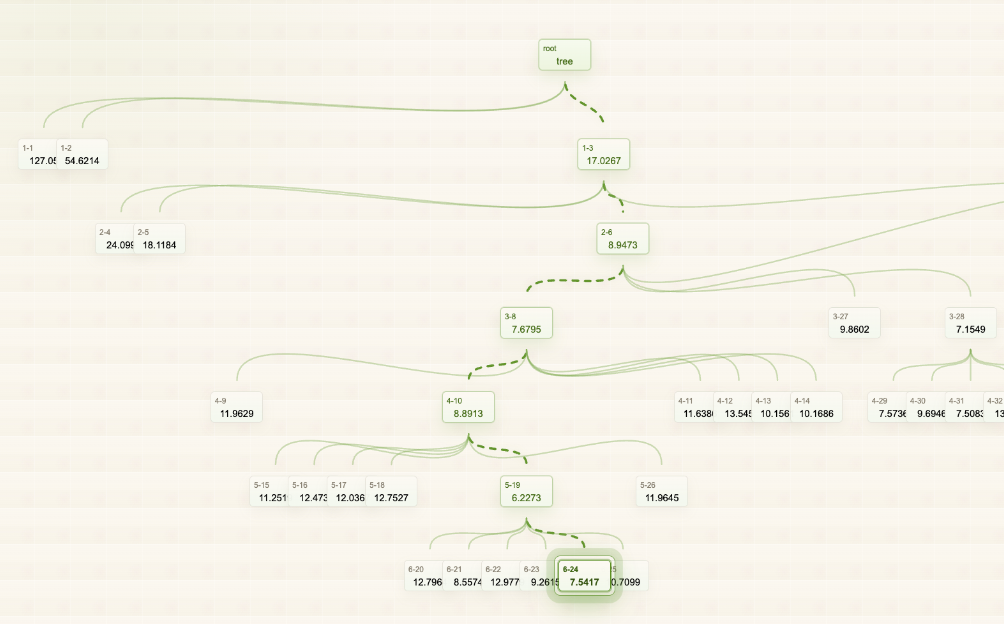}
    \caption{An example of the expansion tree on the Influenza-USA dataset. The number in each node denotes the validation loss. The best node is highlighted in green, and the path from the root node to this node is also highlighted.}
    \label{fig:tree}
\end{figure}

\section{Impact Statement}
This paper proposed a benchmark for evaluating the LLM's capability in generating neural-integrated mechanistic models, and an agentic framework for enabling the LLMs to optimize the neural-integrated mechanistic models. Our main aim is to explore potentially effective methods for generating the neural-integrated mechanistic models.

\end{document}